\documentclass{article}

\PassOptionsToPackage{numbers, compress}{natbib}

 \usepackage[main, final]{neurips_2026}

\usepackage[utf8]{inputenc} 
\usepackage[T1]{fontenc}    
\usepackage{hyperref}       
\usepackage{url}            
\usepackage{booktabs}       
\usepackage{amsfonts}       
\usepackage{nicefrac}       
\usepackage{microtype}      
\usepackage{xcolor}         
\usepackage[dvipsnames]{xcolor}

\usepackage{graphicx}
\usepackage{subcaption}
\usepackage{multirow}
\usepackage{array}
\usepackage{siunitx}
\usepackage{amsmath}
\usepackage{amssymb}
\usepackage{mathtools}
\usepackage{amsthm}
\usepackage{longtable}
\usepackage{tabularx}
\usepackage{makecell}
\usepackage{ragged2e}
\usepackage[capitalize,noabbrev]{cleveref}
\usepackage{mdframed}
\usepackage[table]{xcolor}

\usepackage{wrapfig}

\usepackage{mdframed} 

\newenvironment{qtemplate}
{%
  \begin{mdframed}[
    linewidth=0.5pt,
    innertopmargin=5pt,
    innerbottommargin=5pt,
    innerleftmargin=6pt,
    innerrightmargin=6pt,
    skipabove=6pt,
    skipbelow=6pt
  ]
  \small
}
{%
  \end{mdframed}
}

\title{SceneShifter: Training-free Multi-Scene Temporal Control for Audio-driven Human Animation}

\author{%
  Jianzhi Long$^{1}$ \quad
  Rong-Cheng Tu$^{2*}$ \quad
  Hao Guan$^{2}$ \quad
  Siyuan Liang$^{2}$ \\
  \textbf{Shunyu Liu}$^{2}$ \quad
  \textbf{Xiao Luo}$^{3}$ \quad
  \textbf{Dacheng Tao}$^{2*}$ \\[0.5em]
  $^{1}$The University of Sydney, New South Wales, Australia \\
  $^{2}$Generative AI Lab, College of Computing and Data Science, \\Nanyang Technological University, Singapore, Singapore \\
  $^{3}$University of Wisconsin--Madison, Wisconsin, United States
}

\begin{document}
\raggedbottom

\maketitle

\renewcommand{\thefootnote}{\fnsymbol{footnote}}
\footnotetext[1]{Corresponding authors.}

\begin{abstract}
Although audio-driven human animation has achieved impressive realism, it still lacks effective \textbf{Multi-Scene Temporal Control}: orchestrating background transitions at precise timestamps while preserving a coherent foreground subject. We show that this challenge stems from the diffusion transformer's self-attention mechanism, and addressing it requires two capabilities: (1) \textbf{Semantic-Aware Temporal Decoupling}, which isolates background attention across scenes while retaining cross-scene foreground attention for subject consistency; and (2) \textbf{Foreground Motion Localization}, which accurately tracks the foreground subject across latent frames, especially under large motion. 
To address these challenges, we introduce SceneShifter, a training-free framework for precise multi-scene control in audio-driven human animation. 
SceneShifter guides spatio-temporal self-attention by suppressing cross-scene attention among background tokens while preserving cross-scene attention among foreground tokens. 
To support this guidance under large motion, SceneShifter extracts dynamic foreground masks from the self-attention layers and heads that best capture subject trajectories. We further introduce \textbf{SceneShifterBench}, a benchmark designed to evaluate scene-timing accuracy, foreground preservation, and visual quality under large motion, multiple subjects. Experiments show that SceneShifter achieves frame-level scene timing, strong subject preservation, and high visual quality, outperforming existing baselines.
Code is available at https://github.com/JamesLong199/SceneShifter.

\end{abstract}

\section{Introduction}
\label{sec:intro}

\begin{figure}[h]
  \centering
  \includegraphics[width=\linewidth]{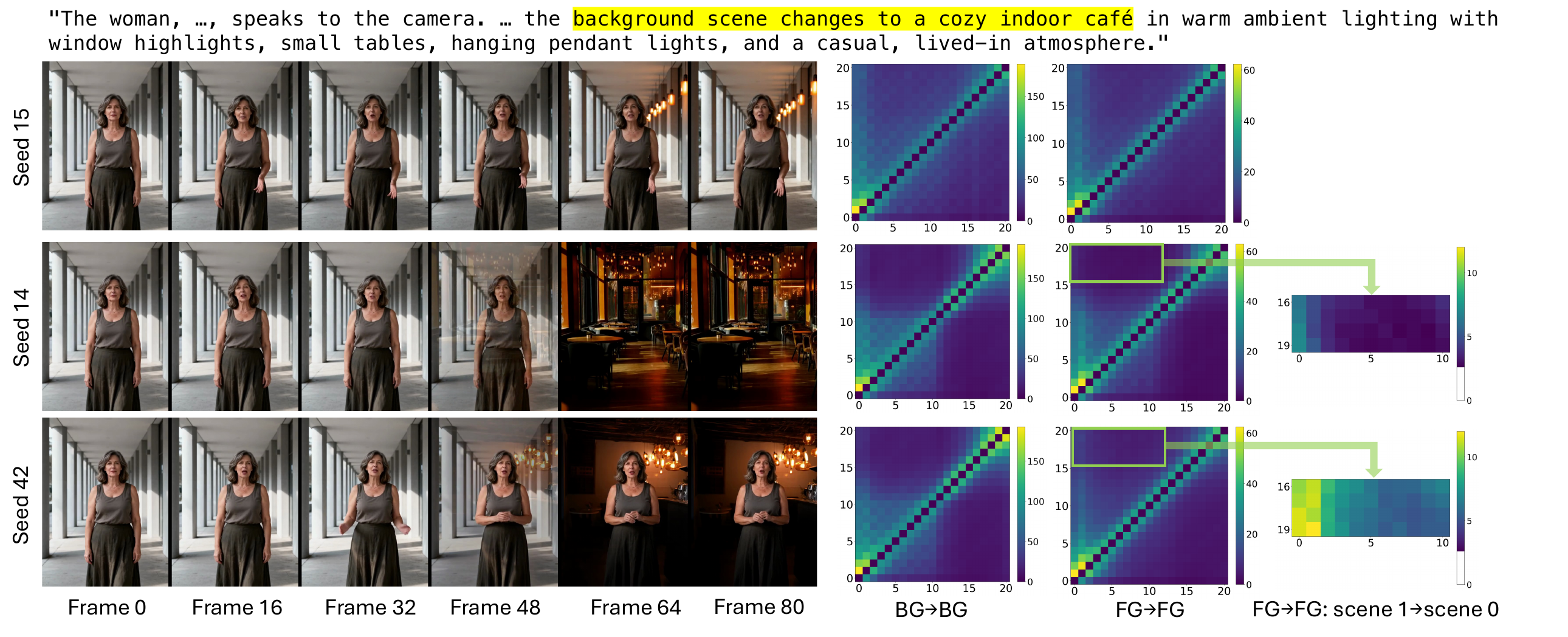}
  \caption{
  Videos generated from the same reference image and multi-scene prompt using different random seeds.
  (\textit{\textbf{Top}}) An example of failing to change the background scene. 
  (\textit{\textbf{Middle}}) An example of failing to preserve the subject across scenes. 
  (\textit{\textbf{Bottom}}) An example of successful generation.
  The results are generated using the prompt shown above with different seeds. 
  }
  \label{fig:text_scene_control}
\end{figure}

\begin{figure}[h]
  \centering
  \begin{subfigure}[t]{0.63\linewidth}
    \centering
    \includegraphics[width=\linewidth]{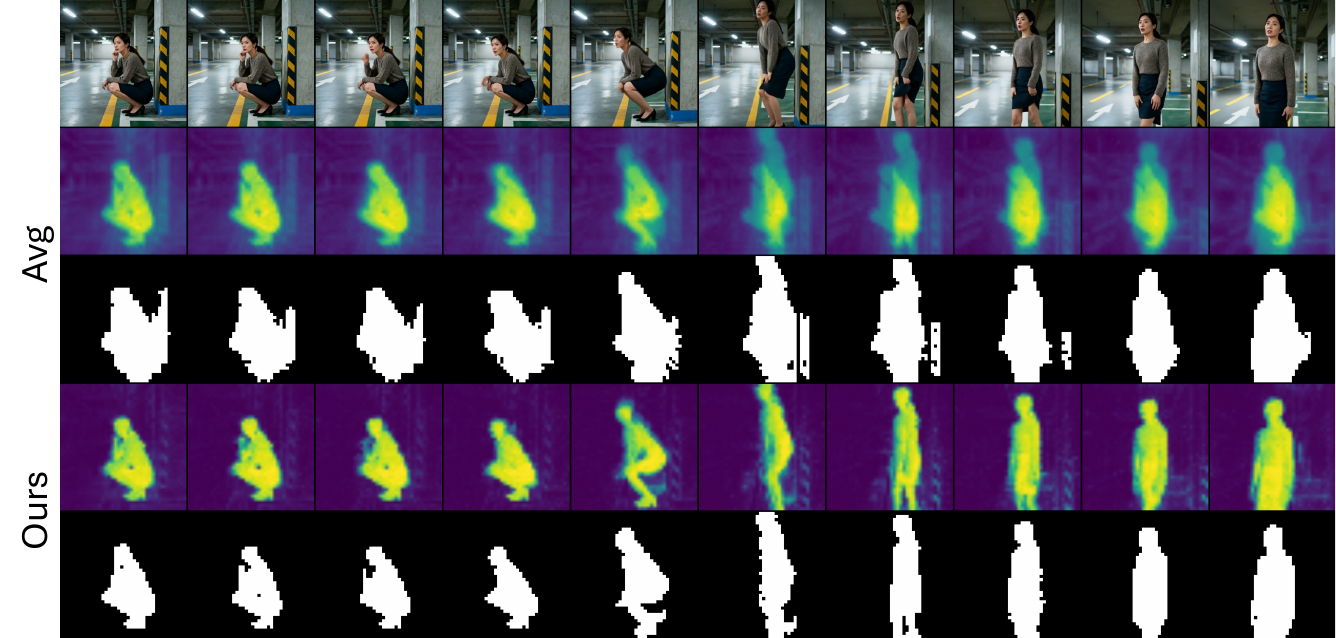}
    \caption{Reference-to-video attention heatmaps and extracted binary masks.
    }
    \label{fig:mask_comparison}
  \end{subfigure}
  \hfill
  \begin{subfigure}[t]{0.35\linewidth}
    \centering
    \includegraphics[width=\linewidth]{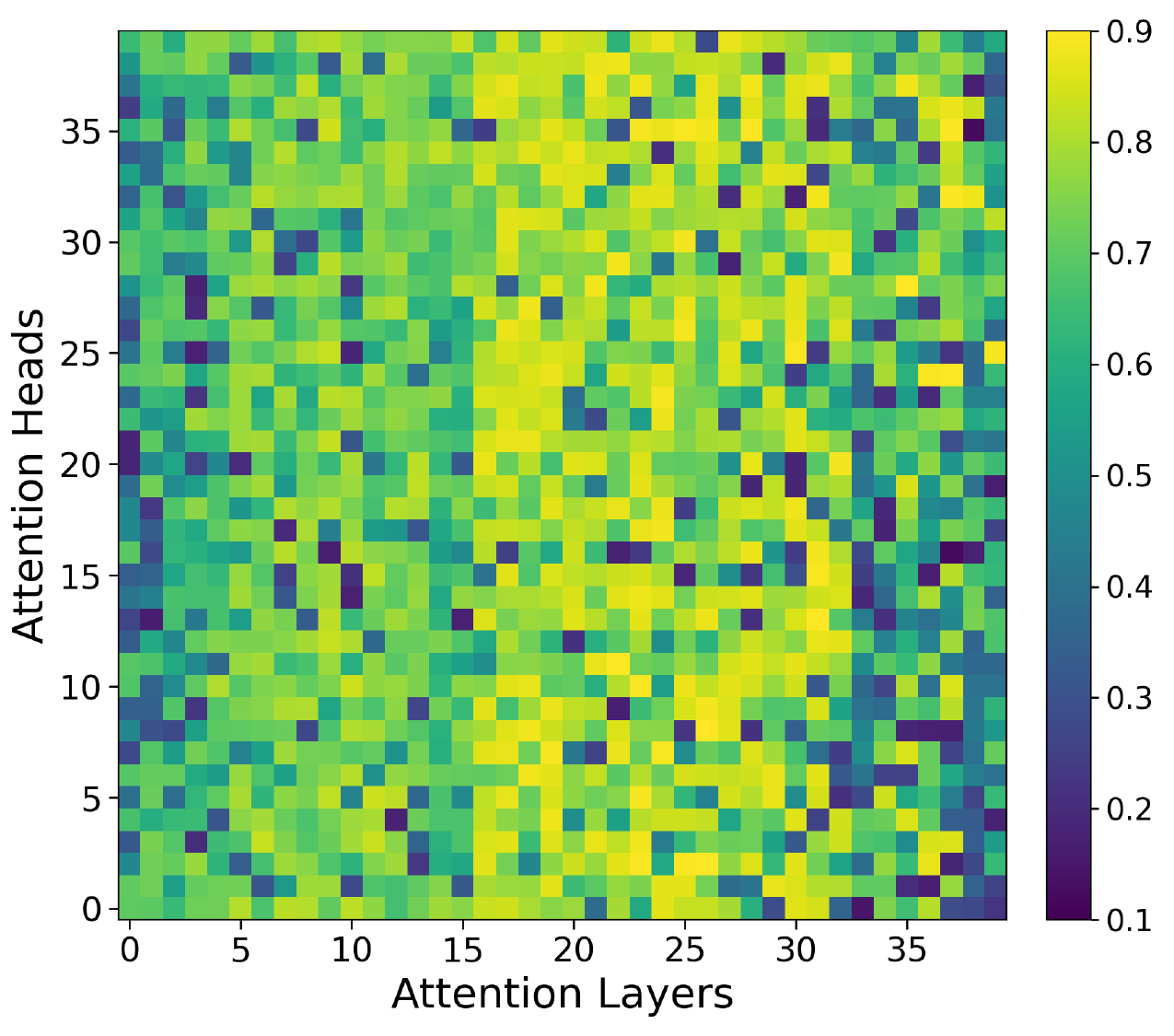}
    \caption{Mask IoU heatmap.}
    \label{fig:mask_iou}
  \end{subfigure}

  \caption{Results of different foreground localization strategies on a video with large subject motion.
  \textbf{Avg}: attention-score averaging across all heads within each DiT layer.}
  \label{fig:ref_attn_mask}
\end{figure}

Generating audio-driven human animation with temporally controllable background scenes is important for scripted video production, virtual presentations, and social media content creation, where scene transitions must align with narration, dialogue, or editing timestamps.
Recent diffusion-based video and audio-driven animation models \cite{wang2025fantasytalking, yang2025infinitetalk, kong2025let, gao2025wan, chen2025hunyuanvideo} have made substantial progress in generating high-quality human videos with synchronized lip motion, expressive body movement, and realistic backgrounds.

However, existing methods struggle with \textbf{Multi-Scene Temporal Control}: generating multiple background scenes at specified timestamps while preserving a coherent foreground subject. 
As shown in Fig.~\ref{fig:text_scene_control}, state-of-the-art image-to-video models \cite{wan2025wan} typically fail in two ways: 
the generated video either remains locked to the original background, or changes the scene at the cost of foreground subject consistency.
Our analysis of inter-frame self-attention shows that successful multi-scene generation requires \textbf{Semantic-Aware Temporal Decoupling}, in which background and foreground tokens follow different temporal attention patterns. 
Background attention should be decoupled across scene segments so that each segment can form a distinct scene at the desired time. 
In contrast, foreground attention should remain connected across scenes so that foreground tokens in the new scene can still attend to those in the previous scene, preserving subject appearance and motion continuity. 
When this cross-scene foreground reference is weak, the subject may change appearance, shift position, or disappear after the transition. These observations motivate explicit self-attention guidance that isolates background attention across scenes while preserving semantic continuity among foreground tokens.

Another challenge is \textbf{Foreground Motion Localization} under large subject motion, because attention guidance relies on separating foreground from background tokens.
As shown in Fig.~\ref{fig:mask_comparison}, large displacement and pose variation make accurate subject localization across frames difficult.
MultiTalk~\cite{kong2025let} uses reference-to-video attention scores to localize subjects across frames by comparing each video token's average similarity to masked subject and background regions in the reference image.
However, the attention scores are averaged across heads in each DiT layer, making this approach unreliable under large motion.
Different layers and heads respond to different visual cues: some attend to the subject's current location, whereas others remain biased toward the reference position or background context. 
Averaging them therefore blurs motion-specific evidence and produces ambiguous masks.
As shown by the IoU heatmap in Fig.~\ref{fig:mask_iou}, masks from individual DiT self-attention heads differ widely in accuracy. 
The resulting masks can cover both the subject's current and reference positions, as illustrated in the second and third rows of Fig.~\ref{fig:mask_comparison}.
In multi-scene generation, these inaccurate masks can cause semantic leakage across scenes and degrade background perceptual quality.


To address these challenges, we introduce \textbf{SceneShifter}, a training-free framework for precise multi-scene temporal control in audio-driven human animation.
SceneShifter begins with spatio-temporal self-attention guidance to achieve semantic-aware temporal decoupling: background tokens are isolated across scene segments to enforce precise temporal cuts, while foreground tokens retain cross-scene attention to preserve subject appearance and motion continuity.
To make this guidance robust under large subject motion, we analyze the sensitivity of self-attention to subject motion trajectories across layers and heads, and propose a fine-grained foreground motion localization method that derives accurate masks from the most informative self-attention components, as illustrated in the last two rows of Fig.~\ref{fig:mask_comparison}.
To enable systematic evaluation, we establish \textbf{SceneShifterBench}, a dedicated benchmark for multi-scene audio-driven human animation that includes challenging cases with large motion and multiple subjects, along with evaluation protocols for temporal accuracy, subject preservation, and visual quality.
Extensive experiments demonstrate that SceneShifter achieves accurate frame-level scene timing, robust subject consistency, and high-quality generation, substantially outperforming existing state-of-the-art methods.

In summary, we introduce \textbf{SceneShifter}, a training-free framework for multi-scene temporal control in audio-driven human animation, with the following contributions:

\begin{itemize}

\item We formulate multi-scene temporal control as a semantic-dependent attention control problem, where background tokens require cross-scene isolation and foreground tokens require cross-scene continuity. Based on this formulation, we propose a spatio-temporal self-attention guidance strategy for precise scene cuts and coherent subject preservation.

\item We analyze the sensitivity of self-attention to subject motion trajectories across layers and heads, and propose a fine-grained \textbf{Foreground Motion Localization} method that extracts dynamic foreground masks from the most informative self-attention components for large-motion scenarios.


\item We establish \textbf{SceneShifterBench}, a dedicated benchmark for evaluating multi-scene audio-driven human animation under large motion, multiple subjects, and occlusions. 
Experiments demonstrate strong overall performance across temporal accuracy, subject preservation, and visual quality.

\end{itemize}

\section{Related Work}




\subsection{Audio-driven Human Animation}

GAN-based~\cite{goodfellow2014generative} audio-driven human animation methods provide coarse lip synchronization with limited realism, expressiveness, and temporal stability~\cite{prajwal2020lip, zhang2021flow, wang2021audio2head, wang2022one, zhang2023sadtalker, long2026simdem}.
Diffusion-based methods improve visual fidelity and motion quality.
Earlier approaches extend latent diffusion models~\cite{rombach2022high} with 3D convolutions, audio attention, and temporal modules~\cite{tian2024emo, zheng2024memo, meng2025echomimicv2, lin2024cyberhost, long2026lightning}, but primarily target frontal, tightly cropped subjects and generalize poorly to complex backgrounds.

More recent approaches build on DiT-based video generators~\cite{wan2025wan, kong2024hunyuanvideo}, enabling audio-driven animation with wider shots and richer backgrounds~\cite{wang2025fantasytalking, yang2025infinitetalk, kong2025let, gao2025wan, chen2025hunyuanvideo}.
These methods add audio conditioning to pretrained backbones but assume a single continuous background and lack precise temporal scene control.
Our work enables training-free, temporally controlled background changes while remaining compatible with these pretrained models.



\subsection{Temporally Controlled Video Generation}

These works study how to control when text-specified events or concepts appear in a video. 
MinT \cite{wu2025mind} controls temporal events through temporal cross-attention and rescaled positional encoding but requires fine-tuning on temporally annotated datasets.
SwitchCraft~\cite{xu2026switchcraft} steers frame-level attention toward relevant prompts without additional training, but its cross-attention control struggles with large changes such as background transitions under complex prompts.
DiTCtrl~\cite{cai2025ditctrl} treats multi-prompt generation as temporal video editing, using training-free attention sharing in MM-DiT architectures~\cite{yang2025cogvideox}.
It produces smooth transitions but assumes fixed-duration scenes and struggles with cross-scene subject preservation.
Our training-free method supports arbitrary scene timing through self-attention decoupling and preserves subject consistency through fine-grained foreground motion localization.


\subsection{Mask Guidance in Attention}
Mask guidance localizes edit regions and constrains attention manipulation in diffusion-based image and video editing~\cite{shuai2024survey,sun2407diffusion}.
Many methods~\cite{hertz2022prompt, qi2023fatezero, cai2025ditctrl} derive masks from target tokens' cross-attention maps, but background leakage and distractor objects can make these masks noisy~\cite{yang2023dynamic}, limiting their reliability in complex, multi-subject scenes.
MultiTalk~\cite{kong2025let} instead uses reference-to-video attention to localize subjects through similarity to masked reference-image regions.
However, masks computed using all attention heads can become ambiguous under large subject motion.
Since our task requires finer foreground tracking for precise scene decoupling and natural subject-scene interaction,
we analyze localization across self-attention layers and heads, and select informative components that accurately track subject trajectories.
\section{Methods}

\begin{figure}[h]
  \centering
  \includegraphics[width=\linewidth]{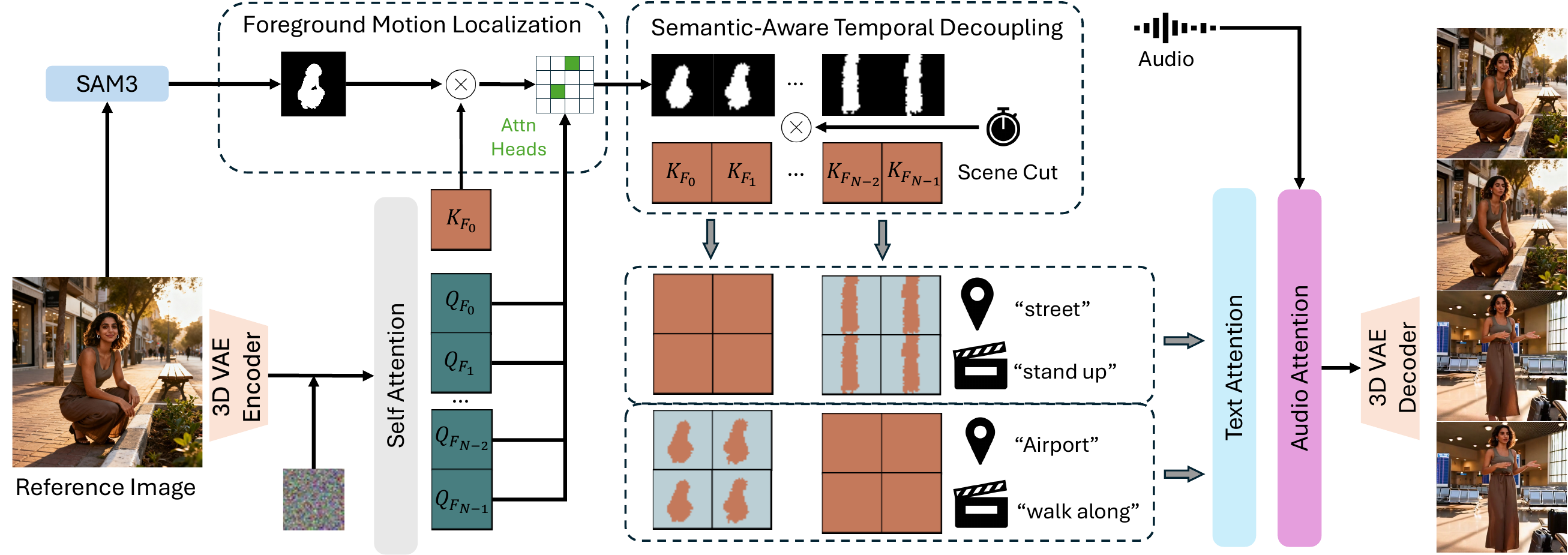}
  \caption{
  Pipeline of the proposed SceneShifter framework. 
  }
  \label{fig:pipeline}
\end{figure}

In this section, we present \textbf{SceneShifter}, a training-free framework for multi-scene temporal control in audio-driven human animation. 
As shown in Fig.~\ref{fig:pipeline}, SceneShifter augments pretrained DiT-based video generators \cite{kong2025let,wang2025fantasytalking} with explicit temporal control, without modifying model parameters. 
Given a reference image, audio, user-specified scene cut timings, and scene-specific text prompts, SceneShifter partitions video generation into scene segments. 
Within each DiT block, it applies self-attention guidance to enforce precise scene transitions while preserving subject consistency across scenes. 
To accurately separate foreground and background, SceneShifter also performs foreground motion localization for fine-grained subject tracking. 
Sec.~\ref{sec:preliminary} introduces the task formulation, Sec.~\ref{sec:semantic_aware_temporal_decoupling} presents semantic-aware temporal decoupling, and Sec.~\ref{sec:foreground_motion_localization} describes foreground motion localization.

\subsection{Preliminaries}
\label{sec:preliminary}

\subsubsection{Audio-driven Human Animation}

Recent audio-driven human animation methods \cite{wang2025fantasytalking, yang2025infinitetalk, kong2025let, gao2025wan} build on powerful I2V diffusion models to generate expressive human motion and realistic backgrounds. 
They extend a DiT-based I2V backbone \cite{wan2025wan} with audio conditioning modules, enabling audio+image-to-video (AI2V) generation.

Given a reference image $I_{\mathrm{ref}}$, speech audio $A$, and a text prompt $T$, AI2V generates a video in which the subject in $I_{\mathrm{ref}}$ speaks according to $A$ and moves within a scene described by $T$. 
At inference, $I_{\mathrm{ref}}$ is encoded by a VAE and concatenated to the noisy video latent $z_t$ at the start of denoising. 
Each DiT block then applies self-attention to model spatiotemporal dependencies, performs text cross-attention with text condition $c_{\mathrm{text}}$ encoded from $T$ \cite{raffel2020exploring}, and applies audio cross-attention with audio condition $c_{\mathrm{audio}}$ extracted from $A$ \cite{baevski2020wav2vec}. 
Together, these attention modules integrate visual, textual, and auditory signals to produce audio-synchronized human animation.

\subsubsection{Multi-Scene Temporal Control}

We define \textbf{Multi-Scene Temporal Control} as changing the background scene at a user-specified time while preserving consistent foreground subject appearance and motion. 
To specify the transition, we introduce a temporal condition signal $f_{\mathrm{cut}}$ that encodes the scene-cut timing in the latent frame space and inject $f_{\mathrm{cut}}$ into the model's attention mechanisms. 
Because the video latents are temporally downsampled, $f_{\mathrm{cut}}$ corresponds to a scalar index compressed from the original video frame space by a factor of 4. 
Although our method naturally extends to videos with more than two scenes, for clarity it is presented in the two-scene setting in the following sections, where the generated video $\mathcal{V}=\{\mathcal{V}_{S_0}, \mathcal{V}_{S_1}\}$ is divided into segments corresponding to the initial scene $S_0$ and the new scene $S_1$.

\subsection{Semantic-Aware Temporal Decoupling}
\label{sec:semantic_aware_temporal_decoupling}


Semantic-aware temporal decoupling requires distinct temporal attention patterns for background (BG) and foreground (FG) tokens.
BG tokens should be isolated across scenes to enforce changes at specified timestamps, while FG tokens should retain cross-scene attention to preserve subject appearance and motion continuity.
We first analyze these patterns in the self-attention maps of a pretrained DiT~\cite{wan2025wan}, and then introduce spatio-temporal self-attention guidance to enforce them.

\paragraph{Self-attention analysis.}

We generate videos from the same reference image and multi-scene prompt using three random seeds.
Fig.~\ref{fig:text_scene_control} shows three outcomes: no background change, background change with subject loss, and background change with subject preservation.
We extract FG masks~\cite{carion2025sam} from the generated frames and downsample them to latent resolution to separate FG and BG tokens.
For seed~14, where the subject disappears, we reuse the last detected FG mask as a pseudo mask for subsequent frames to keep attention statistics comparable across seeds.

We compute inter-frame self-attention statistics over the first 50\% of denoising steps, accumulating BG-to-BG and FG-to-FG scores separately for each pair of latent frames.
The resulting BG$\rightarrow$BG heatmaps in Fig.~\ref{fig:text_scene_control} associate successful background transitions with suppressed cross-scene BG attention.
In the first row, strong cross-frame BG attention supports a single continuous background.
Videos with scene changes instead exhibit clear attention separation at the boundary, supporting temporal BG decoupling to initiate a new background at the specified timestamp.

Foreground attention follows a different temporal pattern.
To examine cross-scene foreground references, we visualize the block where FG tokens in Scene~1 query those in Scene~0, labeled FG$\rightarrow$FG: Scene~1$\rightarrow$Scene~0 in Fig.~\ref{fig:text_scene_control}.
Comparing subject-inconsistent and successful generations shows the importance of retaining attention to foreground tokens in the preceding scene.
Weak or absent cross-scene FG attention can lead to subject disappearance or identity changes after the transition.
These observations motivate semantic-aware temporal decoupling: suppressing cross-scene BG attention while preserving FG references.

\paragraph{Spatio-temporal self-attention guidance.}
We enforce semantic-aware temporal decoupling through spatio-temporal self-attention guidance.
Each DiT block projects its input hidden features $H_t$ into
$Q=H_tW_Q$, $K=H_tW_K$, and $V=H_tW_V$, which are then partitioned at the user-specified cut frame $f_{\mathrm{cut}}$ into $(Q_{S_0},K_{S_0},V_{S_0})$ and $(Q_{S_1},K_{S_1},V_{S_1})$, as illustrated in Fig.~\ref{fig:pipeline}. 
Per-frame foreground masks further separate each scene's keys and values into FG and BG components:
\begin{align}
K_{S_0} &= \{K^{\mathrm{FG}}_{S_0},K^{\mathrm{BG}}_{S_0}\}, \quad
V_{S_0} = \{V^{\mathrm{FG}}_{S_0},V^{\mathrm{BG}}_{S_0}\}, \\
K_{S_1} &= \{K^{\mathrm{FG}}_{S_1},K^{\mathrm{BG}}_{S_1}\}, \quad
V_{S_1} = \{V^{\mathrm{FG}}_{S_1},V^{\mathrm{BG}}_{S_1}\}.
\end{align}
Here, braces denote concatenation along the token dimension.
Each scene retains all within-scene features but excludes BG features from the other scene:
\begin{align}
\omega_{S_0}^{\mathrm{self}}
&= \mathrm{Attn}(Q_{S_0}, \{K_{S_0},K^{\mathrm{FG}}_{S_1}\}, \{V_{S_0},V^{\mathrm{FG}}_{S_1}\}),
\label{eq:scene0_self_attn} \\
\omega_{S_1}^{\mathrm{self}}
&= \mathrm{Attn}(Q_{S_1}, \{K^{\mathrm{FG}}_{S_0},K_{S_1}\}, \{V^{\mathrm{FG}}_{S_0},V_{S_1}\}).
\label{eq:scene1_self_attn}
\end{align}
This isolates backgrounds across scenes while preserving FG references for appearance consistency and motion continuity.
Each scene segment is conditioned on its corresponding prompt through text attention.
Adaptation to MM-DiT architectures is detailed in Appendix~\ref{sec:appendix_backbones}.

\paragraph{Transition context window.}
While cross-scene background suppression enables scene separation, it can disrupt temporal continuity near scene boundaries.
To mitigate this, we introduce a short transition context window: queries in the first $T_{\mathrm{trans}}$ latent frames of each new scene additionally attend to both foreground and background from the last $T_{\mathrm{ctx}}$ latent frames of the preceding scene.
This relaxation is applied only during late denoising steps to smooth transitions while preserving scene separation, with the original attention policy retained elsewhere.

\subsection{Foreground Motion Localization}
\label{sec:foreground_motion_localization}

Semantic-aware temporal decoupling requires accurate FG-BG separation, especially under large subject motion, as shown in the first row of Fig.~\ref{fig:mask_comparison}.
In image-to-video generation, reference-to-video attention can be used to localize the FG subject by measuring each video token's similarity to the FG and BG regions of the reference image~\cite{kong2025let}.
The attention map between video-token queries $Q_{vid}$ and reference-token keys $K_{ref}$ is computed as:
\begin{equation}
P = \mathrm{Softmax}\left(\frac{Q_{vid}(K_{ref})^\top}{\sqrt{d}}\right),
\label{eq:ref_attn}
\end{equation}
where $d$ is the head dimension.
For each video token $i$, we average its attention scores over the reference FG and BG token sets $\Omega_{FG}$ and $\Omega_{BG}$ to obtain the FG and BG scores, respectively:
\begin{equation}
S_{FG}(i)=\frac{1}{|\Omega_{FG}|}\sum_{j\in\Omega_{FG}} P_{ij}, \quad
S_{BG}(i)=\frac{1}{|\Omega_{BG}|}\sum_{j\in\Omega_{BG}} P_{ij}.
\label{eq:fg_similarity}
\end{equation}
We classify token \(i\) as foreground when \(S_{FG}(i)>S_{BG}(i)\), and as background otherwise.


In MultiTalk~\cite{kong2025let}, the attention scores are averaged across all attention heads before computing FG and BG scores.
This can produce ambiguous masks covering both the subject's reference and current positions, as shown in the second and third rows of Fig.~\ref{fig:mask_comparison}.
In multi-scene generation, such masks can allow nearby background objects to leak across scenes.
We attribute this ambiguity to differences in attention head sensitivity to subject motion.
To examine this, we use segmentation masks~\cite{carion2025sam} extracted from the generated frames as ground truth and report each layer-head pair's mean foreground and background intersection over union (IoU) in Fig.~\ref{fig:mask_iou}.
The substantial variation indicates that only a subset of heads reliably tracks subject trajectories.


We select heads with high localization accuracy and average their reference-to-video attention maps before computing FG and BG scores.
For MultiTalk, we generate 20 single-scene videos using reference images from SceneShifterBench and large-motion prompts.
We rank layer-head pairs by their mean foreground and background IoU against these ground-truth masks, averaged across the videos and the first 40\% of denoising steps.
Table~\ref{tab:head_iou} shows that averaging L37H32 and L37H35 outperforms the best individual head.
As shown in the last two rows of Fig.~\ref{fig:mask_comparison}, the resulting masks better follow subject motion.


\begin{table}[h]
\centering
\caption{Localization accuracy of individual and averaged heads.
L and H denote layer and head indices; ``+'' denotes attention-score averaging.}
{\footnotesize
\setlength{\tabcolsep}{4pt}
\resizebox{\linewidth}{!}{%
\begin{tabular}{lccccccc}
\toprule
Head(s)
& L39H25 & L37H32 & L37H35 & L32H3 & L37H24
& L37H32 + L37H35 & L39H25 + L37H35 \\
\midrule
Mean IoU
& 0.862 & 0.860 & 0.859 & 0.858 & 0.855
& \textbf{0.870} & 0.869 \\
\bottomrule
\end{tabular}}
}
\label{tab:head_iou}
\end{table}
\section{Experiments}

\subsection{SceneShifterBench} 
\label{sec:evaluation_benchmark}


Existing human animation datasets~\cite{zhang2021flow,meng2025echomimicv2} mainly feature heads or upper bodies with limited backgrounds.
To address this gap,
we introduce SceneShifterBench, a multimodal benchmark with full-body subjects, diverse backgrounds, and task-specific metrics for multi-scene temporal control.

\paragraph{Dataset.}


SceneShifterBench contains 150 samples, each pairing a reference image with scene-specific prompts and speech audio.
The images are generated by Seedream~4.0~\cite{seedream2025seedream}, covering diverse subject appearances, poses, backgrounds, and aspect ratios, with 40 multi-subject cases.
The prompts in each sample describe up to five scenes with subject motions,
while the audio is drawn from~\cite{zhang2021flow}.

\paragraph{Evaluation Metrics.}

We evaluate temporal accuracy, human quality, and background quality.
Temporal accuracy comprises \textbf{Scene Cut Success Rate (Succ.)}, the percentage of videos with the correct number of detected cuts, and \textbf{Scene Timing Accuracy (Acc.)}, the percentage of expected scene cuts correctly detected within their ground-truth frame intervals.
Human quality includes \textbf{Spatial Consistency}, measured by human-mask IoU across scene boundaries and averaged over videos with valid detected cuts, \textbf{Identity} and \textbf{Clothes} from VBench2~\cite{zheng2025vbench}, and \textbf{Lip Synchronization (Sync.)}~\cite{prajwal2020lip}.
For identity, we report raw feature similarity rather than thresholded success rate.
Background quality comprises VBench \textbf{Imaging Quality} and \textbf{Aesthetic Quality}~\cite{huang2024vbench}, alongside \textbf{Scene Semantics}, evaluated only on newly generated scenes.
Scene semantics sums two normalized VLM scores for background--prompt alignment and subject--environment interaction, yielding a score in $[0,2]$, evaluated using LLaVA-Video-7B~\cite{zhang2024video}.
Detailed protocols are provided in Appendix~\ref{app:background_semantics}.

\subsection{Experiment Settings}

\paragraph{Baselines.}

We evaluate our method on five base models: MultiTalk \cite{kong2025let}, FantasyTalking \cite{wang2025fantasytalking}, InfiniteTalk \cite{yang2025infinitetalk}, WanS2V \cite{gao2025wan}, and HunyuanVideo-Avatar \cite{chen2025hunyuanvideo}.
For comparison, we integrate DiTCtrl \cite{cai2025ditctrl} and SwitchCraft \cite{xu2026switchcraft} into the underlying video diffusion backbones of these models, adapting their attention control mechanisms to the AI2V setting.
DiTCtrl restricts attention to tokens within each scene, except at selected denoising timesteps and layers where KV sharing is enabled.
During KV sharing, FG queries in Scene~1 attend only to FG keys and values in Scene~0, while BG queries in Scene~1 attend only to BG keys and values in Scene~0.
In SwitchCraft, the attention modulation is activated during the first 50\% of denoising steps, following the original paper.
We use scene names as action words and determine event length ratios from the scene lengths in latent frames.


\paragraph{Implementation Details.}

For most backbones, we briefly delay self-attention guidance to preserve foreground motion continuity while enforcing background separation, and enable audio attention only during the latter half of denoising to limit interference with scene changes while maintaining lip synchronization, as detailed lip motion emerges during middle-to-late denoising stages~\cite{peng2025omnisync}.
For foreground motion localization, we use SAM~3 \cite{carion2025sam} to extract the human mask from the reference image and compute reference-to-video attention maps using backbone-specific sets of attention heads.
All experiments are conducted on NVIDIA A800 GPUs with 80\,GB of VRAM.
We apply background-only query scaling adapted from~\cite{jin2023training} to mitigate background degradation in short scenes.
Backbone-specific parameter settings of SceneShifter and baselines are provided in Appendix~\ref{sec:appendix_backbones}.

\subsection{Performance Analysis}

\begin{figure}[h]
  \centering
  \includegraphics[width=\linewidth]{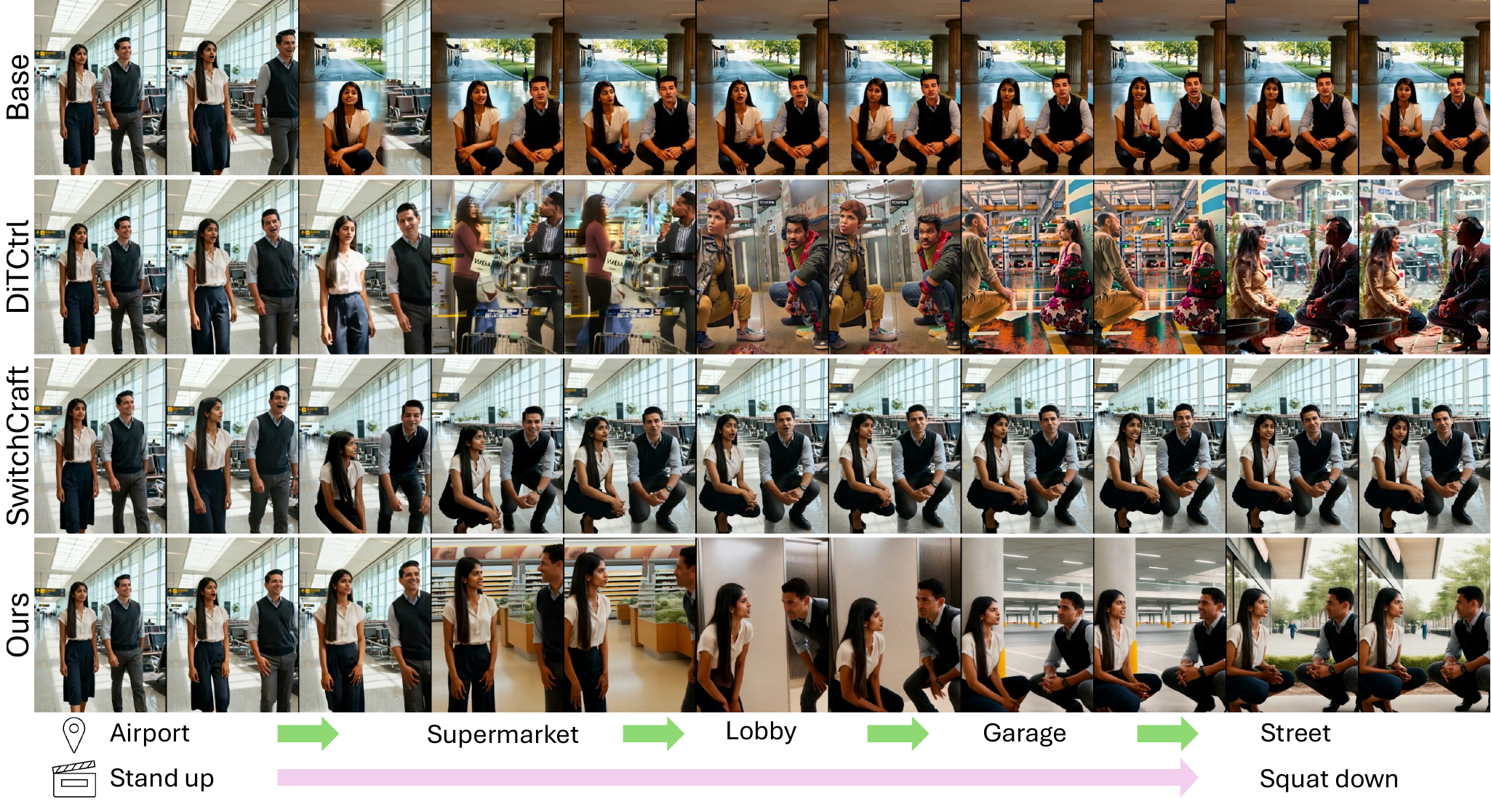}
  \caption{
  Qualitative results generated by the base model, DiTCtrl, SwitchCraft, and our method using the MultiTalk backbone.
  }
  \label{fig:main_result}
\end{figure}

\begin{table}[t]
\centering
\caption{
Quantitative results on the SceneShifterBench, generated using MultiTalk, FantasyTalking, InfiniteTalk, WanS2V, and HunyuanVideo-Avatar backbones.
\textbf{Succ.}: scene cut success rate.
\textbf{Acc.}: scene timing accuracy.
\textbf{Sync.}: lip sync.
Higher is better for all metrics with best scores in \textbf{bold}.
}
{\footnotesize
\setlength{\tabcolsep}{3pt} 
\newcommand{\graycell}[1]{\cellcolor{gray!15}{#1}}
\begin{tabular}{llccccccccc} 

\toprule
\multicolumn{2}{c}{\raisebox{-0.9ex}{Method}}
& \multicolumn{2}{c}{Temporal Accuracy}
& \multicolumn{4}{c}{Human Quality}
& \multicolumn{3}{c}{Background Quality} \\
\cmidrule(lr){3-4} \cmidrule(lr){5-8} \cmidrule(lr){9-11}

&& \textbf{Succ.}
& \textbf{Acc.}
& \textbf{Spatial}
& \textbf{Clothes}
& \textbf{Identity}
& \textbf{Sync.}
& \textbf{Aesthetic}
& \textbf{Imaging}
& \textbf{Semantics} \\

\midrule

\multirow{4}{*}{\rotatebox[origin=c]{90}{MultiTalk}}
& Base
& 38.7\% & 17.5\% & 0.834 & 83.4\% & 0.579 & \textbf{1.46} & 0.541 & 67.5 & 1.61 \\

& DiTCtrl
& 98.7\% & 96.6\% & 0.847 & 31.7\% & 0.407 & 1.23 & 0.523 & 63.3 & 1.55 \\

& SwitchCraft
& 70.0\% & 30.6\% & 0.862 & 78.5\% & 0.577 & 1.36 & \textbf{0.565} & 67.6 & 1.48 \\

& \graycell{Ours}
& \graycell{\textbf{99.3\%}}
& \graycell{\textbf{98.1\%}}
& \graycell{\textbf{0.943}}
& \graycell{\textbf{100.0\%}}
& \graycell{\textbf{0.623}}
& \graycell{1.45}
& \graycell{0.542}
& \graycell{\textbf{71.1}}
& \graycell{\textbf{1.88}} \\

\midrule

\multirow{4}{*}{\rotatebox[origin=c]{90}{\shortstack{Fantasy\\Talking}}}
& Base
& 44.7\% & 17.2\% & 0.770 & 90.3\% & 0.483 & 0.99 & 0.532 & 67.1 & 1.37 \\

& DiTCtrl
& 98.7\% & \textbf{91.9\%} & 0.845 & 31.2\% & 0.359 & \textbf{1.01} & 0.509 & 53.6 & 1.20 \\

& SwitchCraft
& 32.0\% & 21.6\% & 0.575 & 54.9\% & 0.365 & 0.96 & 0.482 & 56.8 & 1.31 \\

& \graycell{Ours}
& \graycell{\textbf{99.3\%}}
& \graycell{89.1\%}
& \graycell{\textbf{0.924}}
& \graycell{\textbf{98.6\%}}
& \graycell{\textbf{0.545}}
& \graycell{0.99}
& \graycell{\textbf{0.541}}
& \graycell{\textbf{70.6}}
& \graycell{\textbf{1.74}} \\

\midrule

\multirow{4}{*}{\rotatebox[origin=c]{90}{\shortstack{Infinite\\Talk}}}
& Base
& 34.0\% & 10.9\% & 0.802 & 77.9\% & 0.517 & 1.33 & 0.553 & 68.1 & 1.67 \\

& DiTCtrl
& \textbf{98.7\%} & \textbf{94.4\%} & 0.853 & 33.6\% & 0.416 & 1.18 & 0.530 & 63.5 & 1.57 \\

& SwitchCraft
& 34.0\% & 13.1\% & 0.802 & 78.4\% & 0.519 & 1.37 & \textbf{0.556} & 67.8 & 1.71 \\

& \graycell{Ours}
& \graycell{98.0\%}
& \graycell{\textbf{94.4\%}}
& \graycell{\textbf{0.948}}
& \graycell{\textbf{100.0\%}}
& \graycell{\textbf{0.628}}
& \graycell{\textbf{1.41}}
& \graycell{0.546}
& \graycell{\textbf{71.7}}
& \graycell{\textbf{1.87}} \\

\midrule

\multirow{4}{*}{\rotatebox[origin=c]{90}{WanS2V}}
& Base
& 1.3\% & 0.6\% & 0.976 & \textbf{100.0\%} & 0.786 & \textbf{1.29} & \textbf{0.629} & \textbf{75.7} & 1.00 \\

& DiTCtrl
& \textbf{100.0\%} & \textbf{100.0\%} & 0.788 & 12.4\% & 0.350 & 1.12 & 0.505 & 63.9 & 0.93 \\

& SwitchCraft
& 0.7\% & 0.0\% & 0.928 & \textbf{100.0\%} & \textbf{0.787} & \textbf{1.29} & 0.620 & 72.9 & 1.00 \\

& \graycell{Ours}
& \graycell{\textbf{100.0\%}}
& \graycell{\textbf{100.0\%}}
& \graycell{\textbf{0.983}}
& \graycell{\textbf{100.0\%}}
& \graycell{0.756}
& \graycell{1.23}
& \graycell{0.605}
& \graycell{74.1}
& \graycell{\textbf{1.62}} \\

\midrule

\multirow{4}{*}{%
  \rotatebox[origin=c]{90}{\shortstack{Hunyuan\\Video-\\Avatar}}}
& Base
& 51.3\% & 17.0\% & 0.776 & 66.7\% & 0.610 & 1.31 & \textbf{0.540} & 68.5 & 1.29 \\

& DiTCtrl
& \textbf{100.0\%} & 82.8\% & 0.817 & 32.8\% & 0.412 & 1.03 & 0.534 & 64.0 & 1.37 \\

& SwitchCraft
& 54.7\% & 18.2\% & 0.735 & 62.1\% & 0.598 & 1.20 & 0.518 & 67.9 & 1.20 \\

& \graycell{Ours}
& \graycell{\textbf{100.0\%}}
& \graycell{\textbf{85.3\%}}
& \graycell{\textbf{0.959}}
& \graycell{\textbf{100.0\%}}
& \graycell{\textbf{0.639}}
& \graycell{\textbf{1.37}}
& \graycell{0.538}
& \graycell{\textbf{68.9}}
& \graycell{\textbf{1.41}} \\

\bottomrule
\end{tabular}
}
\label{tab:main_result}
\end{table}



Table~\ref{tab:main_result} shows that SceneShifter achieves strong overall performance on SceneShifterBench, enabling accurate scene transitions while maintaining subject consistency and visual quality across five backbones.
Note that WanS2V Base fails to generate any new background and all its detected cuts are false positives.
Its high spatial consistency and visual-quality scores arise from maintaining a largely unchanged scene.
SwitchCraft struggles to induce the requested scene changes, potentially reflecting the limited influence of its text cross-attention modulation in the AI2V setting.
DiTCtrl enables scene changes but compromises visual quality, which may stem from excluding within-scene attention during KV sharing.
These results highlight the importance of separating backgrounds across scenes while retaining sufficient attention context for coherent generation.

Fig.~\ref{fig:main_result} illustrates these differences on a challenging multi-subject example involving four scene transitions and a pose change from standing to squatting.
The base model fails to generate the full scene sequence and blends street and lobby elements into a single background, with visible identity drift across the transition.
DiTCtrl produces substantial foreground and background artifacts and fails to preserve subject appearance, while SwitchCraft largely retains the initial airport background.
In contrast, SceneShifter generates all four new scenes with distinct backgrounds while preserving subject appearance and a coherent progression from standing to squatting.

\begin{table}[!ht]
\centering
\caption{Comparison with independent segment generation using MultiTalk on SceneShifterBench.}
\label{tab:independent_pipeline}
{\footnotesize
\setlength{\tabcolsep}{3pt}
\begin{tabular}{lccccccccc}
\toprule
\multirow{2}{*}{Method}
& \multicolumn{2}{c}{Temporal Accuracy}
& \multicolumn{4}{c}{Human Quality}
& \multicolumn{3}{c}{Background Quality} \\
\cmidrule(lr){2-3}
\cmidrule(lr){4-7}
\cmidrule(lr){8-10}
& \textbf{Succ.} & \textbf{Acc.}
& \textbf{Spatial} & \textbf{Clothes} & \textbf{Identity} & \textbf{Sync.}
& \textbf{Aesthetic} & \textbf{Imaging} & \textbf{Semantics} \\
\midrule
Segment-wise
& \textbf{99.3\%} & 96.9\%
& 0.903 & \textbf{100.0\%} & 0.532 & 1.26
& \textbf{0.553} & 70.1 & 1.83 \\
SceneShifter
& \textbf{99.3\%} & \textbf{98.1\%}
& \textbf{0.943} & \textbf{100.0\%}
& \textbf{0.623} & \textbf{1.45}
& 0.542 & \textbf{71.1} & \textbf{1.88} \\
\bottomrule
\end{tabular}
}
\end{table}

\begin{figure}[t]
    \centering
    \includegraphics[width=\textwidth]{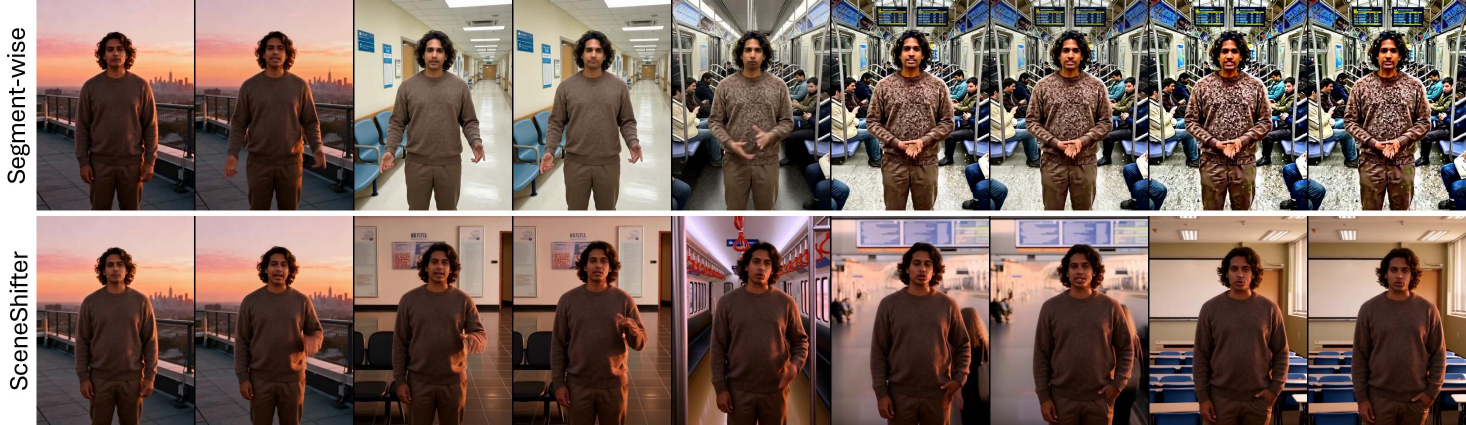}
    \caption{Comparison with independent segment generation and image-based background replacement.
    The prompted scene sequence is a rooftop terrace, a hospital waiting area, a subway train, a train station concourse, and a university classroom.
    }
    \label{fig:qual_pipeline}
\end{figure}

\paragraph{Comparison with Independent Segment Generation.}
To evaluate the advantages of self-attention guidance for multi-scene temporal control, we compare SceneShifter with a sequential pipeline that generates each segment independently.
At each scene boundary, a state-of-the-art open-source image model~\cite{wu2025qwen} replaces the background of the preceding segment's final frame, which then conditions the next segment.
Table~\ref{tab:independent_pipeline} reports lower identity similarity and spatial consistency for this pipeline, suggesting that repeated editing compromises subject appearance and position.
In addition, spatial discontinuities and independent generation from segmented audio may disrupt audiovisual continuity and contribute to lower lip synchronization.
Fig.~\ref{fig:qual_pipeline} provides a qualitative comparison, showing visible subject appearance changes across segments.
The incomplete replacement of the train station with a classroom further illustrates the limited reliability of image-based background editing.
Beyond these quality limitations, the sequential pipeline requires an additional image-editing operation at every scene boundary.
The image-editing model also introduces GPU memory and model-loading requirements, complicating deployment.

\subsection{Ablation Study}


\paragraph{Foreground motion localization.}
We compare our selected-head localization with a fixed foreground mask from the reference image and masks estimated by averaging reference-to-video attention across all heads.
Table~\ref{tab:ablation_results} shows that our method achieves the best temporal accuracy and lip synchronization, with competitive identity, spatial consistency, and background semantics.
Fixed masks cannot follow subject motion, disrupting cross-scene foreground references and weakening identity and spatial consistency.
All-head averaging slightly improves subject consistency, but its broader, ambiguous masks allow background leakage and reduce semantic accuracy.

\begin{figure}[h]
  \centering
  \includegraphics[width=\linewidth]{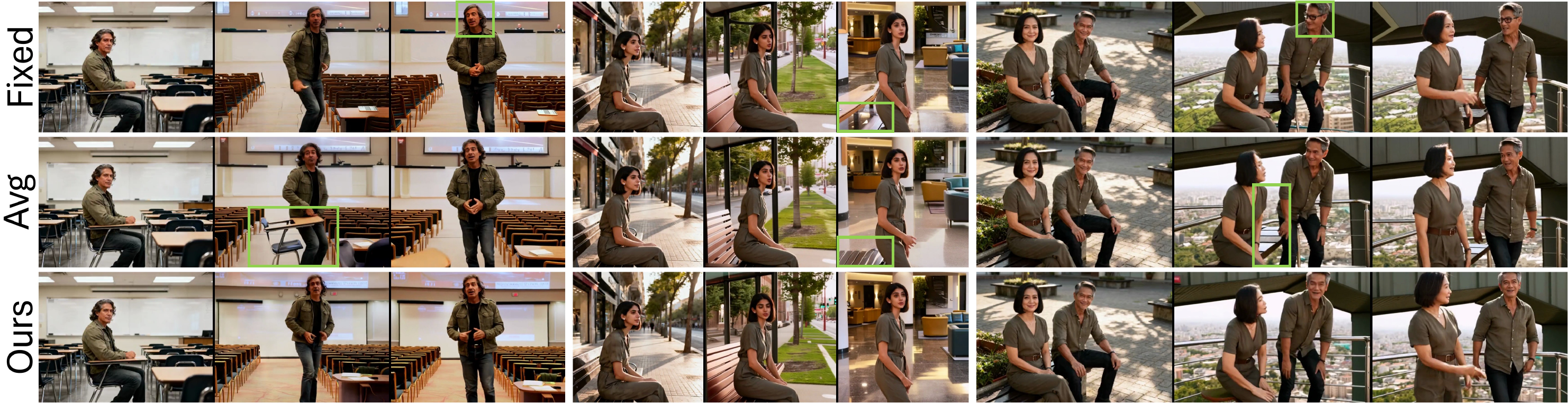}
  \caption{
  Comparison of foreground localization mask strategies. \textbf{Fixed}: fixed mask from reference image. \textbf{Avg}: mask computed from reference-to-video scores averaged across all attention heads. 
  }
  \label{fig:ablation_result}
\end{figure}

Fig.~\ref{fig:ablation_result} compares three challenging cases with substantial subject motion.
Fixed masks cause appearance changes in the first and third examples, altering the hairstyle and adding glasses, respectively.
Both baseline mask strategies also cause background leakage.
In the second example, the subject leaves the fixed mask region, allowing background tokens corresponding to the bench to serve as foreground references.
All-head averaging produces masks that also cover the subject's reference position, causing seat leakage in the first two examples and fence discontinuities in the third.
In comparison, our foreground localization better follows subject motion, preserving subject appearance while reducing background leakage.

\begin{table}[h]
\centering
\caption{Foreground localization ablation using MultiTalk on all 150 samples of SceneShifterBench. The transition context window is enabled for all strategies.}
\label{tab:ablation_results}
{\footnotesize
\setlength{\tabcolsep}{3pt}
\begin{tabular}{lccccccccc}
\toprule
\multirow{2}{*}{Method}
& \multicolumn{2}{c}{Temporal Accuracy}
& \multicolumn{4}{c}{Human Quality}
& \multicolumn{3}{c}{Background Quality} \\
\cmidrule(lr){2-3}
\cmidrule(lr){4-7}
\cmidrule(lr){8-10}
& \textbf{Succ.} & \textbf{Acc.}
& \textbf{Spatial} & \textbf{Clothes} & \textbf{Identity} & \textbf{Sync.}
& \textbf{Aesthetic} & \textbf{Imaging} & \textbf{Semantics} \\
\midrule
Fixed
& 98.7\% & 95.0\%
& 0.933 & \textbf{100.0\%} & 0.608 & 1.41
& 0.543 & 71.1 & \textbf{1.91} \\
Avg
& 98.0\% & 96.3\%
& \textbf{0.947} & \textbf{100.0\%}
& \textbf{0.625} & 1.39
& \textbf{0.545} & \textbf{71.3} & 1.85 \\
Ours
& \textbf{99.3\%} & \textbf{98.1\%}
& 0.943 & \textbf{100.0\%} & 0.623
& \textbf{1.45}
& 0.542 & 71.1 & 1.88 \\
\bottomrule
\end{tabular}
}
\end{table}



\begin{figure}[h]
    \centering
    \includegraphics[width=\linewidth]{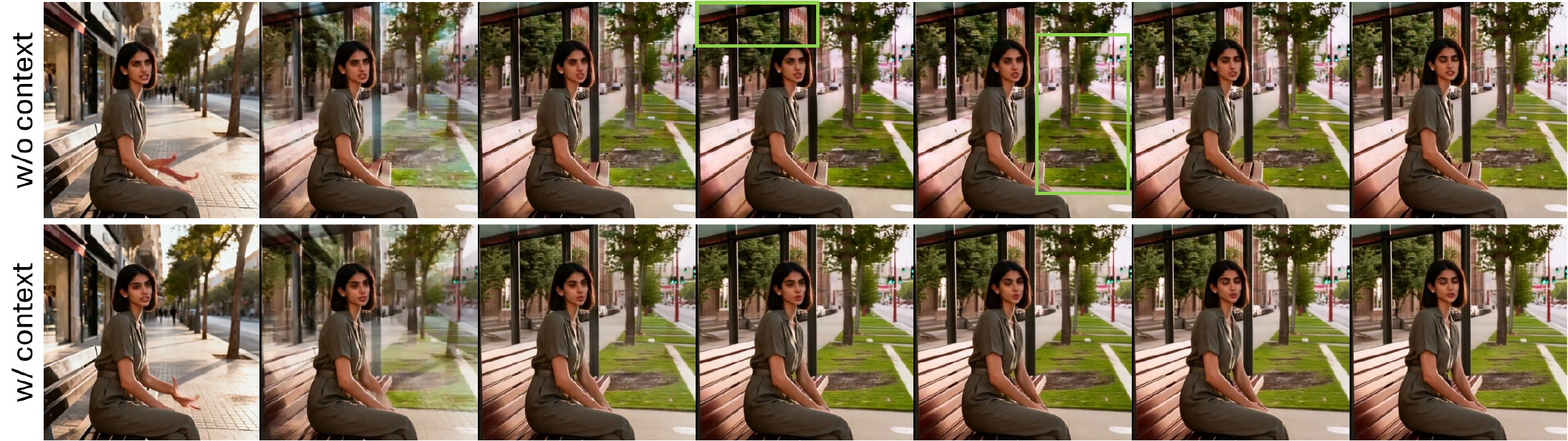}
    \caption{Transition context window ablation on MultiTalk.
    Frames around an urban-street-to-bus-stop transition are shown in temporal order.
    }
    \label{fig:qual_transition}
\end{figure}

\paragraph{Transition context window.}
We ablate the transition context window on MultiTalk to examine its effect on visual continuity near scene boundaries.
As shown in Fig.~\ref{fig:qual_transition}, background structures fluctuate across nearby frames without the window, as highlighted by the green boxes.
Providing context from the preceding scene reduces these artifacts while preserving subject appearance and distinct backgrounds.

\subsection{User Study}
\label{sec:user_study}

\begin{wraptable}{r}{0.53\textwidth}
\centering
\caption{User preference shares (\%). Higher is better.}
\label{tab:user_study}
\small
\setlength{\tabcolsep}{3pt}
\begin{tabular}{@{}lcccc@{}}
\toprule
Criterion & Base & DiTCtrl & SwitchCraft & Ours \\
\midrule
Transition quality    & 23.6 & 5.0 & 16.4 & \textbf{55.0} \\
Subject consistency   & 13.6 & 3.6 & 17.1 & \textbf{65.7} \\
Background quality   & 25.7 & 3.6 & 18.6 & \textbf{52.1} \\
\bottomrule
\end{tabular}
\end{wraptable}



We compare the base model, DiTCtrl, SwitchCraft, and SceneShifter on MultiTalk using 10 randomly selected examples.
Method names are concealed and result order is randomized per question.
All 14 participants select the best result for each example on transition quality, subject appearance and motion consistency, and background quality, yielding 140 judgments per criterion.
SceneShifter leads all three criteria in Table~\ref{tab:user_study}.

\section{Limitations}
\label{sec:limitation}


As a training-free method, SceneShifter is subject to the capability limits of the underlying pretrained models~\cite{tu2026spagent} and cannot induce large movements or pose changes that they fail to generate.
Despite query scaling~\cite{jin2023training}, background quality can degrade when individual scenes are much shorter than the base model's training clips.
In rare cases of multi-subject video generation, severe inter-subject occlusion can merge foreground masks, causing subject disappearance and appearance drift upon re-entry.
Illustrative examples of these limitations are provided in Appendix~\ref{sec:appendix_failures}.



\section{Conclusion}

We present \textbf{SceneShifter}, a training-free framework for precise multi-scene temporal control in audio-driven human animation.
Through \textbf{Semantic-Aware Temporal Decoupling}, it isolates background attention across scenes for accurate scene cuts and semantics while preserving cross-scene foreground attention for subject appearance and motion consistency.
To handle large subject motion, \textbf{Foreground Motion Localization} selects informative self-attention components to extract dynamic foreground masks.
We also introduce \textbf{SceneShifterBench}, which includes challenging large-motion and multi-subject cases.
Experiments and ablations show that SceneShifter outperforms existing baselines in temporal accuracy, subject preservation, and visual quality.
Future work could integrate SceneShifter into agentic video generation and editing workflows~\cite{tu2026spagent,yao2025survey} to support more flexible and creative applications.

\section{Acknowledgements}

This research / project is supported by the National Research Foundation, Singapore, and Cyber Security Agency of Singapore under its National Cybersecurity R\&D Programme and CyberSG R\&D Cyber Research Programme Office. Any opinions, findings and conclusions or recommendations expressed in these materials are those of the author(s) and do not reflect the views of National Research Foundation, Singapore, Cyber Security Agency of Singapore as well as CyberSG R\&D Programme Office, Singapore.

\bibliographystyle{plainnat}
\bibliography{reference}

\appendix


\clearpage
\appendix
\section{SceneShifterBench Details}
\label{sec:appendix_benchmark}

\subsection{Reference Images and Scene Prompts}
SceneShifterBench contains 150 samples, each comprising a reference image, scene-specific text prompts, speech audio, and prescribed scene-transition timings. We construct image-generation prompts by sampling subject count, appearance, pose, background, and aspect ratio, and synthesize the reference images using Seedream 4.0 \cite{seedream2025seedream}. Each image is paired with speech audio from HDTF \cite{zhang2021flow}. The reference images include 110 single-subject and 40 two-subject examples, covering 190 subjects in total. Subject attributes span nine ethnicity categories, five age groups, and five body types. Both the reference images and scene prompts cover 25 background types, including indoor and outdoor environments.

Table~\ref{tab:benchmark_summary} summarizes the benchmark composition, while Tables~\ref{tab:subject_distribution} and~\ref{tab:scene_distribution} provide detailed distributions. Subject attributes describe the categories specified during synthetic image construction. Image-level statistics use 150 reference images as the denominator; subject-level statistics use 190 subjects.

\begin{table}[h]
\centering
\begin{minipage}[t]{0.57\textwidth}
\vspace{0pt}
\centering
\caption{Composition of SceneShifterBench. Scene counts include the initial scene.}
\label{tab:benchmark_summary}
{\small
\setlength{\tabcolsep}{3pt}
\begin{tabularx}{\linewidth}{@{}>{\raggedright\arraybackslash}Xr@{}}
\toprule
Attribute & Count \\
\midrule
\multicolumn{2}{@{}l}{\textbf{Reference images}} \\
Total images & 150 \\
Single-subject / two-subject & 110 / 40 \\
Total subjects & 190 \\
Ethnicity / age / body-type categories & 9 / 5 / 5 \\
Background types & 25 \\
Aspect ratios & 5 \\
\midrule
\multicolumn{2}{@{}l}{\textbf{Scene prompts}} \\
Two / three / four / five scenes & 50 / 50 / 30 / 20 \\
Total scene instances & 470 \\
Background types & 25 \\
\bottomrule
\end{tabularx}
}

\par\bigskip

\caption{Subject attributes and aspect ratios. Counts cover 190 subjects or 150 images for aspect ratios.}
\label{tab:subject_distribution}
{\small
\setlength{\tabcolsep}{3pt}
\begin{tabularx}{\linewidth}{
    @{}>{\raggedright\arraybackslash}p{0.23\linewidth}
    >{\raggedright\arraybackslash}X@{}
}
\toprule
Attribute & Category (count) \\
\midrule
Gender
& Man (95), woman (95) \\
Ethnicity
& South Asian (24), Hispanic/Latino (23),
Indigenous Australian (23), East Asian (22),
mixed-ethnicity (21), Caucasian (21),
Middle Eastern (21), Southeast Asian (19),
Black (16) \\
Age
& 20s (47), 30s (52), 40s (31),
50s (42), 60s+ (18) \\
Body type
& Average (42), slim (40), curvy (38),
stocky (37), athletic (33) \\
Reference pose
& Sitting (88), standing (76), squatting (26) \\
Aspect ratio
& 4:3 (35), 2:3 (30), 1:1 (29),
3:2 (28), 3:4 (28) \\
\bottomrule
\end{tabularx}
}
\end{minipage}
\hfill
\begin{minipage}[t]{0.40\textwidth}
\vspace{0pt}
\centering
\caption{Background distribution in reference images and prompted scenes, including initial scenes.}
\label{tab:scene_distribution}
{\small
\setlength{\tabcolsep}{3pt}
\begin{tabularx}{\linewidth}{
    @{}>{\raggedright\arraybackslash}Xrr@{}
}
\toprule
Background
& \makecell[r]{Reference\\images}
& \makecell[r]{Prompted\\scenes} \\
\midrule
Covered walkway & 8 & 19 \\
Museum & 8 & 17 \\
Park & 7 & 20 \\
Bookstore & 7 & 17 \\
Rooftop terrace & 7 & 21 \\
Supermarket & 7 & 18 \\
Library & 7 & 17 \\
Urban street & 7 & 20 \\
Elevator lobby & 7 & 17 \\
Train station & 6 & 18 \\
Hotel lobby & 6 & 18 \\
Conference hall & 6 & 20 \\
Parking garage & 6 & 20 \\
Courtyard & 6 & 17 \\
Bus stop & 6 & 22 \\
Beach promenade & 6 & 20 \\
Caf\'e & 6 & 21 \\
Hospital waiting area & 5 & 19 \\
Airport & 5 & 19 \\
Classroom & 5 & 18 \\
Workshop & 5 & 21 \\
Office & 5 & 18 \\
Recording studio & 4 & 19 \\
Podcast room & 4 & 17 \\
Subway & 4 & 17 \\
\midrule
Total & 150 & 470 \\
\bottomrule
\end{tabularx}
}
\end{minipage}
\end{table}

\subsection{Evaluation Metrics}
\label{sec:appendix_metrics}

This section details the evaluation metrics introduced in Sec.~\ref{sec:evaluation_benchmark}.
We report Succ.\ and Acc.\ as percentages, and Spatial Consistency and Identity as raw scores.
Higher values indicate better performance for all four metrics.


\paragraph{Scene Cut Success Rate.}
Let $\mathcal{V}$ denote the evaluated videos with valid scene-transition metadata.

For video $v$, let $K_v$ be the expected number of cuts and
$\hat{\mathbf{c}}_v=(\hat c_{v,1},\ldots,\hat c_{v,\hat K_v})$
the sorted scene-cut indices returned by our PySceneDetect-based pipeline~\cite{pyscenedetect}.
A video is successful when the detected and expected cut counts match:
\begin{equation}
u_v=\mathbf{1}[\hat K_v=K_v],
\qquad
\mathrm{Succ.}
=
\frac{100}{|\mathcal{V}|}
\sum_{v\in\mathcal{V}}u_v.
\end{equation}
A failed detection is treated as producing zero cuts.
This metric measures cut-count agreement, independently of timing, and remains subject to false-positive and missed detections.

\paragraph{Scene Timing Accuracy.}
Scene Timing Accuracy measures the percentage of all expected cuts that are correctly timed in videos with matching detected and expected cut counts.
Detected cuts and expected transitions are paired in chronological order.
Let $I_{v,j}$ denote the accepted video-frame interval for expected transition $j$.

For frame-split evaluation, let $s_{v,j}$ be its latent-frame split index.
With temporal compression ratio $R=4$, the inclusive interval is
\begin{equation}
I_{v,j}=[Rs_{v,j}-(R-1),\,Rs_{v,j}].
\end{equation}

For single-prompt evaluation, expected transitions are extracted in prompt order and assigned to early, middle, or late temporal buckets.
In our experiments, Wan-based models~\cite{kong2025let,wang2025fantasytalking,yang2025infinitetalk,gao2025wan} generate 81-frame videos, while HunyuanVideo-Avatar~\cite{chen2025hunyuanvideo} generates 129-frame videos.
The corresponding inclusive intervals are:
\begin{equation}
\begin{array}{c|ccc}
\text{Video length} & \mathrm{early} & \mathrm{middle} & \mathrm{late}\\
\hline
81  & [0,26] & [27,53] & [54,80]\\
129 & [0,42] & [43,85] & [86,128]
\end{array}.
\end{equation}
Thus, the two conditioning settings use different timing tolerances but share the same per-cut scoring rule.

The number of correctly timed cuts in video $v$ is
\begin{equation}
a_v=
\begin{cases}
\displaystyle
\sum_{j=1}^{K_v}
\mathbf{1}[\hat c_{v,j}\in I_{v,j}],
& \hat K_v=K_v,\\
0, & \text{otherwise}.
\end{cases}
\end{equation}
The dataset-level score is
\begin{equation}
\mathrm{Acc.}
=
100\,
\frac{\sum_{v\in\mathcal{V}}a_v}
{\sum_{v\in\mathcal{V}}K_v}.
\end{equation}
Videos with missing or extra detected cuts receive zero timing credit, while their expected cuts remain in the denominator.
For videos with matching counts, each correctly timed cut contributes independently.

\paragraph{Spatial Consistency.}
Spatial Consistency measures subject-position continuity around detected scene cuts.
We use the cuts returned by the hybrid detector, regardless of whether their count or timing matches the ground truth.
Videos with unsuccessful detector outputs and cuts at or beyond the last video frame are excluded.
Let $M_v(t)\in\{0,1\}^{H\times W}$ denote the human mask extracted using SAM~3~\cite{carion2025sam} at frame $t$.
For multi-subject videos, we extract each subject's mask separately and take their union to obtain $M_v(t)$.

For two binary masks $M$ and $N$, we define
\begin{equation}
\mathrm{IoU}(M,N)=
\begin{cases}
\displaystyle\frac{|M\cap N|}{|M\cup N|},
& |M\cup N|>0,\\
1, & |M\cup N|=0.
\end{cases}
\end{equation}
If no person is detected in either frame of a compared pair, its IoU score is set to zero.
For each detected cut $\hat c_{v,k}$, we use a window radius of $w=1$ and compare adjacent-frame masks.
Let $\mathcal{T}_{v,k}$ contain the frame indices within this window for which both $t$ and $t-1$ are valid:
\begin{equation}
\mathcal{T}_{v,k}
=
\left\{
t\in\mathbb{Z}:
\hat c_{v,k}-1\le t\le\hat c_{v,k}+1,\;
t,t-1\text{ are valid frame indices}
\right\}.
\end{equation}
The per-cut score is
\begin{equation}
S_{v,k}
=
\frac{1}{|\mathcal{T}_{v,k}|}
\sum_{t\in\mathcal{T}_{v,k}}
\mathrm{IoU}\big(M_v(t),M_v(t-1)\big).
\end{equation}
Cuts with no valid frame pairs are excluded.
Let $\mathcal{K}_v$ denote the remaining cuts in video $v$, and $\mathcal{V}_{\mathrm{sp}}$ the videos with at least one valid cut score.
We average over cuts within each video and then equally over eligible videos:
\begin{equation}
\mathrm{Spatial}
=
\frac{1}{|\mathcal{V}_{\mathrm{sp}}|}
\sum_{v\in\mathcal{V}_{\mathrm{sp}}}
\frac{1}{|\mathcal{K}_v|}
\sum_{k\in\mathcal{K}_v}S_{v,k}.
\end{equation}

\paragraph{Identity.}
We evaluate identity preservation using ArcFace~\cite{deng2019arcface}.
Unlike the thresholded success rate in VBench2~\cite{zheng2025vbench}, we report raw cosine similarity.
The input reference image is resized to match the first generated frame when their dimensions differ.
For each evaluated frame, we retain up to five detected faces and select the assignment maximizing similarity to the reference subjects.

Let $f_{v,r}^{\mathrm{ref}}$ denote the reference embedding for subject $r$ and $f_{v,r}(t)$ its assigned embedding in generated frame $t$.
For the set $\mathcal{F}_{v,r}$ of frames with valid assignments, the subject-level score is
\begin{equation}
\mathrm{ID}_{v,r}
=
\frac{1}{|\mathcal{F}_{v,r}|}
\sum_{t\in\mathcal{F}_{v,r}}
\frac{
f_{v,r}(t)^\top f_{v,r}^{\mathrm{ref}}
}{
\|f_{v,r}(t)\|_2
\|f_{v,r}^{\mathrm{ref}}\|_2+\epsilon
},
\qquad \epsilon=10^{-12}.
\end{equation}
A subject requires at least 20 valid frames to be included.
For two-subject videos, the score averages the two subjects' identity similarities.
The reported Identity score is averaged over videos with valid scores.

\subsection{Scene Semantics}
\label{app:background_semantics}

Scene Semantics evaluates background--prompt alignment and subject--environment interaction using LLaVA-Video-7B~\cite{zhang2024video}, leveraging the semantic understanding and reasoning capabilities of Vision-Language Models (VLM)~\cite{tu2025automatic, lan2025survey}.
We construct a video question-answering questionnaire from each scene's ground-truth prompt.
Evaluation includes only videos whose detected cut count matches the expected count, with scene boundaries defined by the detected cuts.
We exclude the opening scene and evaluate only the newly generated scenes.
The reported score sums the two component scores, each in $[0,1]$, yielding a total in $[0,2]$.

\paragraph{Background--prompt alignment.}
For each evaluated scene, we parse its prompt into at most five statements describing the background location and objects.
The location statement takes the form ``The background location is \texttt{<location>}.'' 
Additional statements take the form ``The background contains \texttt{<object>}.'' and are extracted from background objects listed after the location clause.
Object phrases are retained only if they contain a word from a fixed place-and-object vocabulary, such as bench, chair, table, window, floor, or street.
If no usable statement is extracted, we use ``The background scene matches the described setting.''
The VLM judges whether each statement is true, considering only the environment.

\paragraph{Subject--environment interaction.}
We first ask the VLM to classify the subject's pose as standing, sitting, or squatting, then evaluate pose-specific physical plausibility.
Prompts describing walking are assigned the standing category.
Standing and squatting each have one follow-up plausibility question.
Sitting has two sequential checks: whether the subject sits on the expected support and whether the body is naturally integrated with that support.
If the predicted pose does not match the prompted pose, all follow-up checks receive zero.
For sitting, a failed support check also assigns zero to the integration check.
Questions assigned zero by these rules are not sent to the VLM.
For multi-action videos, interaction is evaluated only on the last evaluated scene.

Pose questions require a single-word answer, while the remaining questions require yes/no answers.
Where requested, a negative answer includes an explanation of at most one sentence.
These explanations do not contribute to the score.

\paragraph{Questionnaire templates.}
The following templates specify the evaluated questions.

\noindent\textbf{Background--prompt alignment.}
For each parsed statement \texttt{<gt\_item>}, we ask:
\begin{qtemplate}
Please answer yes or no only for the following question.
IMPORTANT: Focus only on the environment/background.
Ignore the person/subject, faces, clothing, and actions.
Is this statement true in the background? `\texttt{<gt\_item>}'.
Answer with yes or no only.
\end{qtemplate}

\noindent\textbf{Subject--environment interaction.}
Here, \texttt{<subject>} is the subject phrase extracted from the prompt, such as ``the woman'' or ``the man and the woman.''

\noindent\emph{Pose classification:}
\begin{qtemplate}
What pose is \texttt{<subject>} performing?
Choose exactly one word: standing, sitting, or squatting.
\end{qtemplate}

\noindent\emph{Standing plausibility:}
\begin{qtemplate}
Does the standing by \texttt{<subject>} look physically plausible
(feet grounded, no obvious floating, no severe body--environment intersection)?
Answer yes/no.
If no, briefly explain ($\leq 1$ sentence).
\end{qtemplate}

\noindent\emph{Squatting plausibility:}
\begin{qtemplate}
Does the squatting by \texttt{<subject>} look physically plausible and stably grounded
(feet contact the ground plausibly; no obvious hovering, sliding, or severe ground intersection)?
Answer yes/no.
If no, briefly explain ($\leq 1$ sentence).
\end{qtemplate}

\noindent\emph{Sitting support:}
\begin{qtemplate}
Is \texttt{<subject>} sitting on a `\texttt{<expected\_support>}'
(as intended by the prompt)?
Answer yes or no.
If no, briefly explain ($\leq 1$ sentence).
\end{qtemplate}

\noindent\emph{Sitting integration:}
\begin{qtemplate}
Does the sitting by \texttt{<subject>} look naturally supported by the `\texttt{<expected\_support>}'
(hips/thighs contact the seat surface; body not floating above or sinking into it)?
Answer yes/no.
If no, briefly explain ($\leq 1$ sentence).
\end{qtemplate}

If the prompt specifies no support object, the sitting questions refer to a plausible seat or surface.

\paragraph{Example: standing.}
For video~001, the evaluated scene is an indoor recording studio.

\noindent\emph{Scene prompt:}
\begin{qtemplate}
The man and the woman walk along the studio floor in indoor recording studio, soundproof walls, studio monitors, and a microphone stand near the subject, speaking to the camera.
\end{qtemplate}

\noindent\emph{Background--prompt alignment:}
\begin{qtemplate}
Is this statement true in the background? `The background location is indoor recording studio.'
Answer with yes or no only.
\end{qtemplate}

\noindent\emph{Subject--environment interaction:}
\begin{qtemplate}
What pose are the man and the woman performing?
Choose exactly one word: standing, sitting, or squatting.
\end{qtemplate}

\begin{qtemplate}
Does the standing by the man and the woman look physically plausible
(feet grounded, no obvious floating, no severe body--environment intersection)?
Answer yes/no.
If no, briefly explain ($\leq 1$ sentence).
\end{qtemplate}

\paragraph{Example: sitting.}
For video~003, the evaluated scene is an indoor podcast room.

\noindent\emph{Scene prompt:}
\begin{qtemplate}
The woman continues to speak to the camera while sitting on studio chair at the recording table in indoor podcast room, acoustic panels, a small recording table, and neatly arranged audio gear.
\end{qtemplate}

\noindent\emph{Background--prompt alignment:}
\begin{qtemplate}
Is this statement true in the background? `The background location is indoor podcast room.'
Answer with yes or no only.
\end{qtemplate}

\noindent\emph{Subject--environment interaction:}
\begin{qtemplate}
What pose is the woman performing?
Choose exactly one word: standing, sitting, or squatting.
\end{qtemplate}

\begin{qtemplate}
Is the woman sitting on a `chair'?
Answer yes or no.
If no, briefly explain ($\leq 1$ sentence).
\end{qtemplate}

\begin{qtemplate}
Does the sitting by the woman look naturally supported by the `chair'?
Answer yes/no.
If no, briefly explain ($\leq 1$ sentence).
\end{qtemplate}

\paragraph{Component scores.}
For an evaluated scene $\sigma$ with $N_\sigma$ background statements, let $y_{\sigma,i}\in\{0,1\}$ indicate whether the VLM judges statement $i$ to be true.
The background--prompt alignment score is
\begin{equation}
S_{\mathrm{align}}(\sigma)
=
\frac{1}{N_\sigma}
\sum_{i=1}^{N_\sigma}y_{\sigma,i}.
\end{equation}

For interaction, let $p_\sigma$ and $p_\sigma^\star$ denote the predicted and expected poses, respectively, and define
$c_{\mathrm{pose}}(\sigma)=\mathbf{1}[p_\sigma=p_\sigma^\star]$.
For standing or squatting, let $c_{\mathrm{plaus}}(\sigma)$ be the binary outcome of the plausibility check.
For sitting, let $c_{\mathrm{sup}}(\sigma)$ and $c_{\mathrm{integ}}(\sigma)$ be the support and integration outcomes.
The interaction score is
\begin{equation}
S_{\mathrm{int}}(\sigma)
=
\begin{cases}
\displaystyle
\frac{c_{\mathrm{pose}}(\sigma)+c_{\mathrm{plaus}}(\sigma)}{2},
& p_\sigma^\star\in\{\mathrm{standing},\mathrm{squatting}\},\\[6pt]
\displaystyle
\frac{c_{\mathrm{pose}}(\sigma)+c_{\mathrm{sup}}(\sigma)+c_{\mathrm{integ}}(\sigma)}{3},
& p_\sigma^\star=\mathrm{sitting}.
\end{cases}
\end{equation}
All follow-up outcomes are zero when $c_{\mathrm{pose}}(\sigma)=0$.
For sitting, $c_{\mathrm{integ}}(\sigma)=0$ whenever $c_{\mathrm{sup}}(\sigma)=0$.
Skipped checks remain in the denominator.

\paragraph{Aggregation.}
Let $\mathcal{G}_v$ denote the evaluated scenes in video $v$, and
$\mathcal{G}_v^{\mathrm{int}}\subseteq\mathcal{G}_v$ the scenes evaluated for interaction.
For multi-action videos, $\mathcal{G}_v^{\mathrm{int}}$ contains only the last evaluated scene.
We compute the per-video component scores as
\begin{align}
\mu_{\mathrm{align}}(v)
&=
\frac{1}{|\mathcal{G}_v|}
\sum_{\sigma\in\mathcal{G}_v}S_{\mathrm{align}}(\sigma),\\
\mu_{\mathrm{int}}(v)
&=
\frac{1}{|\mathcal{G}_v^{\mathrm{int}}|}
\sum_{\sigma\in\mathcal{G}_v^{\mathrm{int}}}S_{\mathrm{int}}(\sigma).
\end{align}
Let $\mathcal{V}_{\mathrm{sem}}$ denote the eligible evaluated videos.
Each component is averaged equally across videos:
\begin{align}
\mu_{\mathrm{align}}
&=
\frac{1}{|\mathcal{V}_{\mathrm{sem}}|}
\sum_{v\in\mathcal{V}_{\mathrm{sem}}}\mu_{\mathrm{align}}(v),\\
\mu_{\mathrm{int}}
&=
\frac{1}{|\mathcal{V}_{\mathrm{sem}}|}
\sum_{v\in\mathcal{V}_{\mathrm{sem}}}\mu_{\mathrm{int}}(v).
\end{align}
The reported Scene Semantics score is
\begin{equation}
\mathrm{Semantics}
=
\mu_{\mathrm{align}}+\mu_{\mathrm{int}}
\in[0,2].
\end{equation}

\section{Experimental Settings}
\label{sec:appendix_settings}



\subsection{Backbone-Specific Settings}
\label{sec:appendix_backbones}
We implement SceneShifter on MultiTalk~\cite{kong2025let}, FantasyTalking~\cite{wang2025fantasytalking}, InfiniteTalk~\cite{yang2025infinitetalk}, WanS2V~\cite{gao2025wan}, and HunyuanVideo-Avatar~\cite{chen2025hunyuanvideo}.
Table~\ref{tab:backbone_settings} summarizes the sampling settings, attention schedules, foreground localization heads, and transition context windows for each backbone.
We average reference-to-video attention scores over the listed heads for mask extraction and update the masks during the specified denoising steps.
Self-attention guidance, audio conditioning, and transition smoothing follow backbone-specific schedules.
All step indices are zero-based and follow denoising execution order; ranges include both endpoints.
Experiments are conducted on NVIDIA A800 GPUs with 80\,GB of VRAM.

\begin{table}[h]
\centering\small
\caption{Backbone-specific settings. Denoising steps are indexed from zero in execution order, and ranges are inclusive. Transition and context lengths are measured in latent frames.}
\label{tab:backbone_settings}
\resizebox{\textwidth}{!}{%
\begin{tabular}{lccccc}
\toprule
Setting & MultiTalk & FantasyTalking & InfiniteTalk & WanS2V & HunyuanVideo-Avatar \\
\midrule
Denoising steps & 40 & 30 & 40 & 40 & 50 \\
Output frames & 81 & 81 & 81 & 81 & 129 \\
Self-attention guidance start step & 3 & 2 & 3 & 3 & 0 \\
Audio attention start step & 20 & 15 & 20 & 0 & 0 \\
Localization heads (layer, head) & (37,32), (37,35) & (37,32), (25,26), (39,25) & (37,32), (37,35) & (37,32), (37,35) & (19,22) \\
Mask update steps & 0--14 & 0--8 & 0--14 & 0--2 & 0--14 \\
Transition length $T_{\mathrm{trans}}$ & 2 & 2 & 2 & 2 & 2 \\
Context length $T_{\mathrm{ctx}}$ & 1 & 1 & 1 & 1 & 2 \\
Transition window activation steps & 30--39 & 22--29 & 30--39 & 30--39 & 20--49 \\
\bottomrule
\end{tabular}}
\end{table}

\paragraph{Adaptation to HunyuanVideo-Avatar.}
For the MM-DiT architecture in HunyuanVideo-Avatar, we apply semantic-aware temporal decoupling and foreground localization to vision-to-vision attention.
Each scene prompt is independently encoded using the model's text encoder, and the resulting tokens are concatenated into the text stream.
To prevent cross-scene semantic leakage, video and text queries associated with one scene cannot attend to text keys from other scenes.

\paragraph{Baseline Settings.}
For each backbone, the base model, SwitchCraft~\cite{xu2026switchcraft}, and DiTCtrl~\cite{cai2025ditctrl} use the same audio attention start step as SceneShifter (Table~\ref{tab:backbone_settings}).
DiTCtrl activates KV sharing only in the final 20\% of DiT layers during an early denoising window: steps 5--14 for MultiTalk, InfiniteTalk, and WanS2V; 2--14 for FantasyTalking; and 0--14 for HunyuanVideo-Avatar.
All other layers and steps retain within-scene attention.

\section{Additional Quantitative Results}
\label{sec:appendix_quantitative}

\subsection{Robustness Across Random Seeds}
\label{sec:appendix_seeds}
We use seed 43 for all samples in the main experiments.
To assess sensitivity to random initialization, we repeat the evaluation on the full SceneShifterBench using MultiTalk~\cite{kong2025let} with seeds 28 and 56, keeping all other settings fixed.
Table~\ref{tab:random_seeds} reports individual scores and their mean and sample standard deviation across seeds.
Performance remains stable across temporal accuracy, human quality, and background quality, with seed 43 showing no consistent advantage.
These results suggest that SceneShifter's performance is not dependent on a particularly favorable random seed.


\begin{table}[h]
\centering
\small
\setlength{\tabcolsep}{4pt}
\caption{Random seed sensitivity on SceneShifterBench. We report means and sample standard deviations across three seed-level aggregate scores.}
\label{tab:random_seeds}
\resizebox{\textwidth}{!}{%
\begin{tabular}{lccccccccc}
\toprule
\multirow{2}{*}{Seed}
& \multicolumn{2}{c}{Temporal Accuracy}
& \multicolumn{4}{c}{Human Quality}
& \multicolumn{3}{c}{Background Quality} \\
\cmidrule(lr){2-3}
\cmidrule(lr){4-7}
\cmidrule(lr){8-10}
& \textbf{Succ. (\%)} & \textbf{Acc. (\%)}
& \textbf{Spatial} & \textbf{Clothes (\%)} & \textbf{Identity} & \textbf{Sync.}
& \textbf{Aesthetic} & \textbf{Imaging} & \textbf{Semantics} \\
\midrule
43 & 99.3 & 98.1 & 0.943 & 100.0 & 0.623 & 1.45 & 0.542 & 71.1 & 1.88 \\
28 & 100.0 & 97.8 & 0.947 & 100.0 & 0.620 & 1.48 & 0.546 & 68.4 & 1.82 \\
56 & 99.3 & 96.6 & 0.946 & 100.0 & 0.612 & 1.48 & 0.542 & 70.3 & 1.87 \\
\midrule
Mean $\pm$ Std
& $99.6\pm0.4$ & $97.5\pm0.8$
& $0.945\pm0.002$ & $100.0\pm0.0$ & $0.618\pm0.006$ & $1.47\pm0.02$
& $0.543\pm0.002$ & $69.9\pm1.4$ & $1.86\pm0.03$ \\
\bottomrule
\end{tabular}}
\end{table}

\section{Additional Qualitative Results}
\label{sec:appendix_qualitative}

\subsection{Results Across AI2V Backbones}
\begin{figure}[t]
    \centering
    \includegraphics[width=\textwidth]{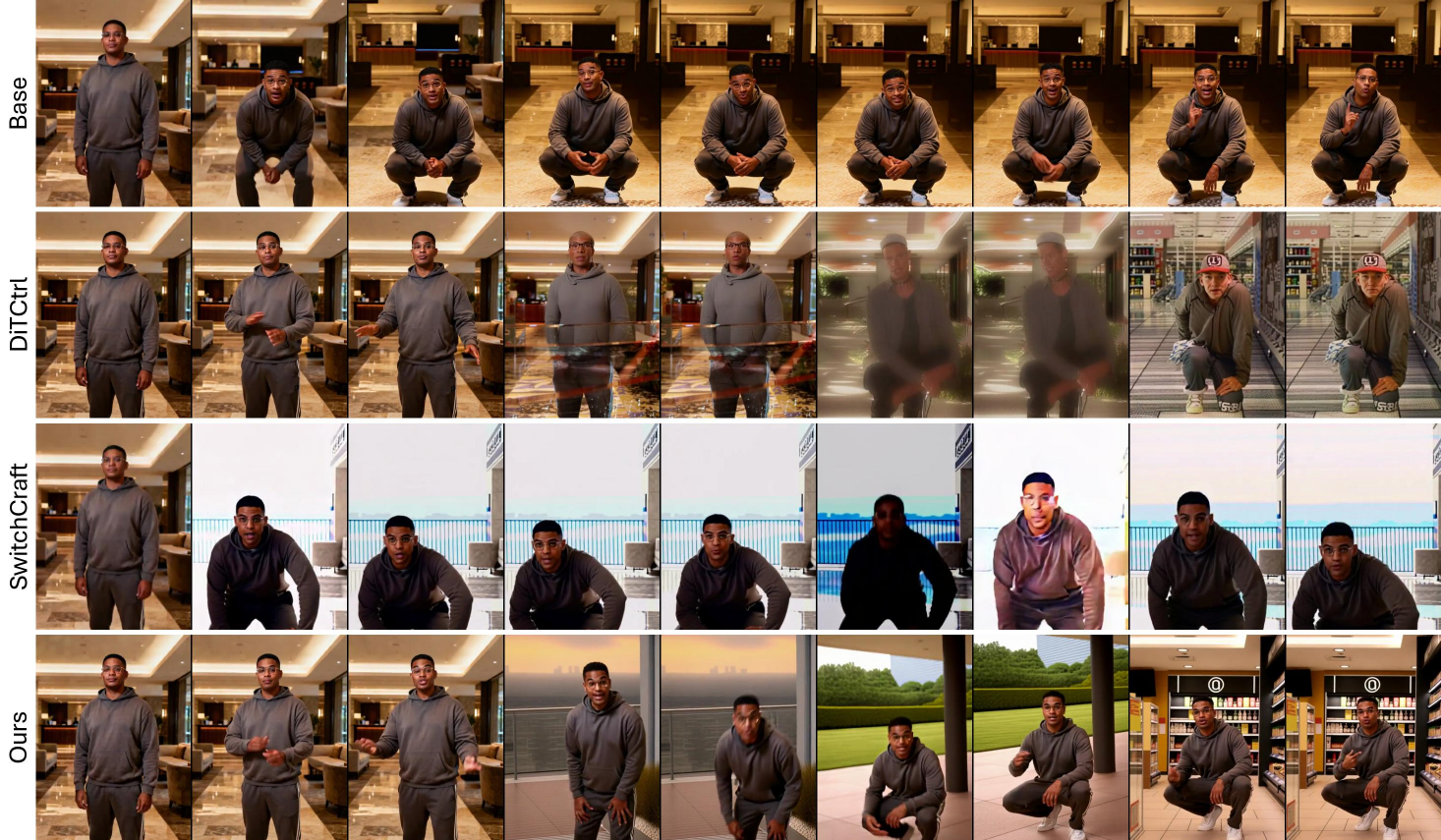}
    \caption{Qualitative comparison on FantasyTalking.
    The initial prompt specifies a squatting action; subsequent prompts specify a rooftop terrace, a covered outdoor walkway, and an indoor supermarket.
    }
    \label{fig:qual_fantasytalking}
\end{figure}

\begin{figure}[t]
    \centering
    \includegraphics[width=\textwidth]{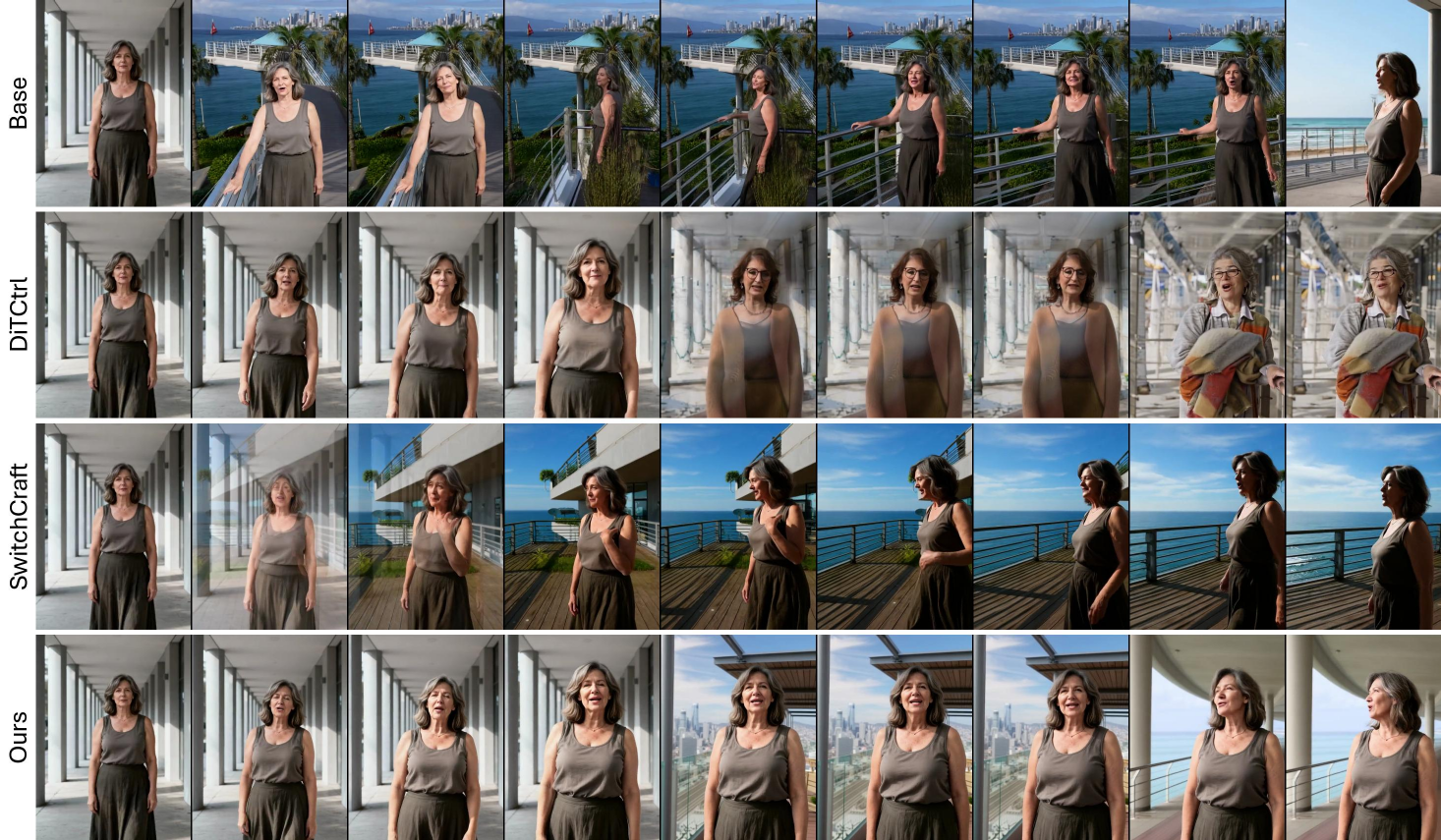}
    \caption{Qualitative comparison on InfiniteTalk.
    The prompted scene sequence is a covered outdoor walkway, a rooftop terrace, and a beachside promenade.
    }
    \label{fig:qual_infinitetalk}
\end{figure}

\begin{figure}[t]
    \centering
    \includegraphics[width=\textwidth]{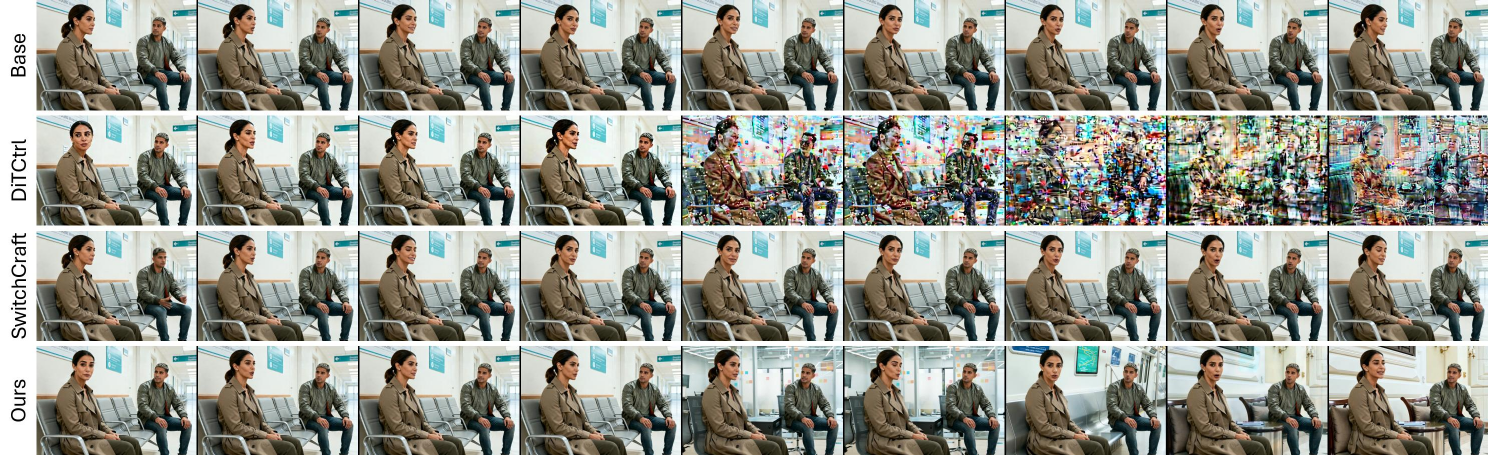}
    \caption{Qualitative comparison on WanS2V.
    The prompted scene sequence is a hospital waiting area, an office, a subway train, and a hotel lobby lounge.
    }
    \label{fig:qual_wans2v}
\end{figure}

\begin{figure}[t]
    \centering
    \includegraphics[width=\textwidth]{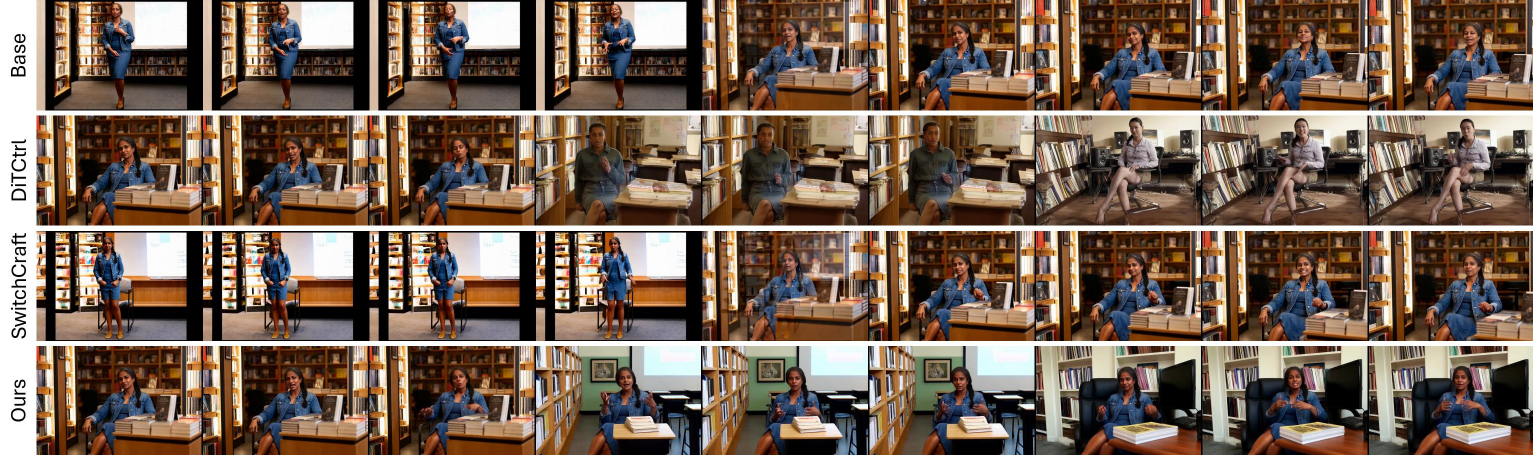}
    \caption{Qualitative comparison on HunyuanVideo-Avatar.
    The prompted scene sequence is a bookstore, a university classroom, and a recording studio.
    }
    \label{fig:qual_hunyuan}
\end{figure}

Figs.~\ref{fig:qual_fantasytalking}--\ref{fig:qual_hunyuan} compare SceneShifter with the base model, DiTCtrl~\cite{cai2025ditctrl}, and SwitchCraft~\cite{xu2026switchcraft} on four additional AI2V backbones, including FantasyTalking~\cite{wang2025fantasytalking}, InfiniteTalk~\cite{yang2025infinitetalk}, WanS2V~\cite{gao2025wan}, and HunyuanVideo-Avatar~\cite{chen2025hunyuanvideo}.
The base models and SwitchCraft often retain the initial background or fail to distinguish successive scenes, while DiTCtrl introduces appearance changes or substantial visual artifacts.
SceneShifter produces more distinct scene sequences while better preserving the foreground subjects.

\subsection{Extensions to Other Generation Settings}
\begin{figure}[!ht]
    \centering
    \includegraphics[width=\textwidth]{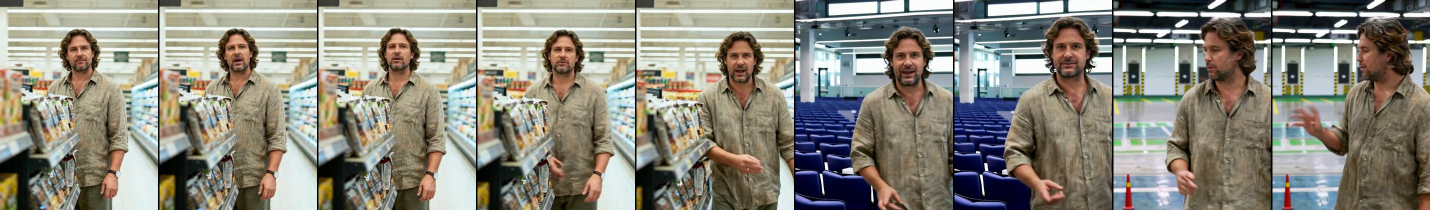}
    \caption{Video-driven animation with SceneShifter on Wan-Animate.
    The prompted scene sequence is a supermarket, a conference hall, and a parking garage.
    Background changes are accompanied by subject gestures controlled by the driving video.}
    \label{fig:qual_wan_animate}
\end{figure}

\paragraph{Video-driven animation.}
Fig.~\ref{fig:qual_wan_animate} demonstrates SceneShifter on Wan-Animate~\cite{cheng2025wan}, where subject motion is controlled by a driving video.
The subject remains visually consistent across changing backgrounds while performing different gestures, illustrating compatibility with video-driven human animation.

\begin{figure}[!ht]
    \centering
    \includegraphics[width=\textwidth]{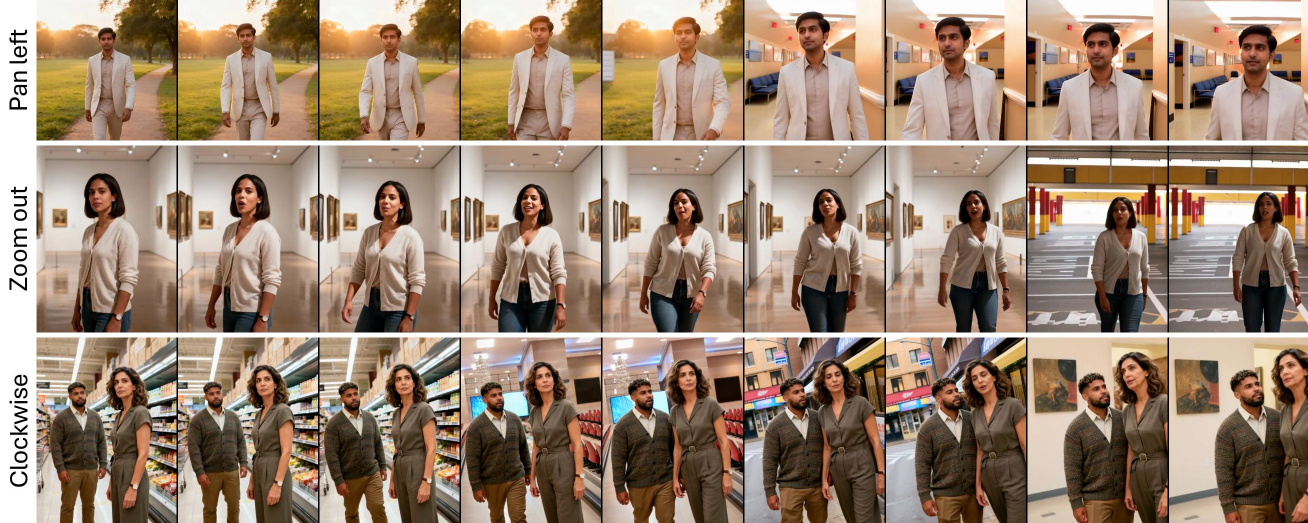}
    \caption{SceneShifter with camera control on Wan2.1-Fun-Camera-Control.
    Top: pan left, with a prompted transition from a park to a hospital waiting area.
    Middle: zoom out, from a museum gallery to a parking garage.
    Bottom: clockwise camera motion, with a prompted sequence of a supermarket, a conference hall, an urban street, and a museum gallery.}
    \label{fig:qual_fun_camera}
\end{figure}

\paragraph{Camera-controlled generation.}
Fig.~\ref{fig:qual_fun_camera} demonstrates SceneShifter on Wan2.1-Fun-Camera-Control~\cite{aigc_apps_VideoX_Fun_2026} with pan-left, zoom-out, and clockwise camera trajectories.
These examples illustrate how SceneShifter can be combined with complementary controls to enable more creative video generation.

\begin{figure}[!ht]
    \centering
    \includegraphics[width=\textwidth]{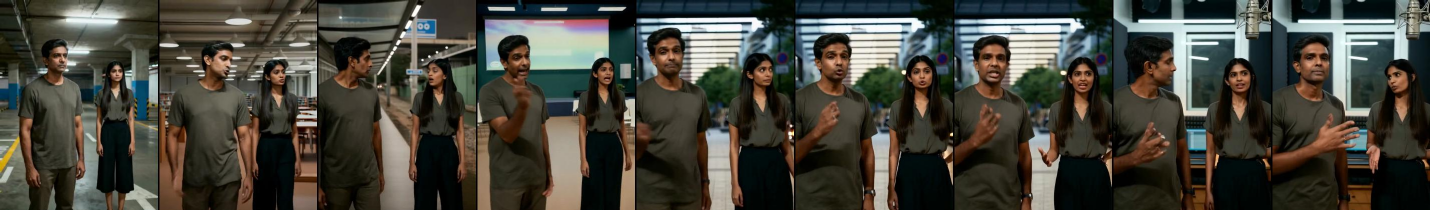}
    \caption{Longer video generation with SceneShifter on MultiTalk.
    The prompted scene sequence is a parking garage, a library study area, a bus stop, a conference hall, an urban street.
    Selected frames span three generated clips and show preservation of both subjects across multiple scene changes.}
    \label{fig:qual_long_video}
\end{figure}

\paragraph{Longer videos.}
Fig.~\ref{fig:qual_long_video} presents a longer 3-clip video generated using MultiTalk~\cite{kong2025let} as backbone, with five motion frames used between consecutive clips.
The example maintains two recognizable subjects through multiple background changes, extending the qualitative evaluation beyond a single short clip.

\section{Limitations}
\label{sec:appendix_failures}

\subsection{Limitations of Base-Model Motion Generation}
\begin{figure}[!ht]
    \centering
    \includegraphics[width=\textwidth]{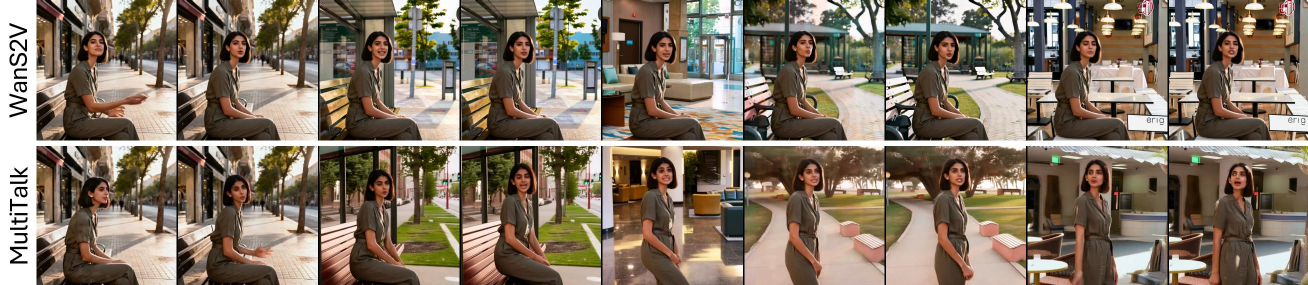}
    \caption{Dependence on the base model's motion-generation capability.
    Both rows apply SceneShifter to the prompted sequence of an urban street, a bus stop, a hotel lobby lounge, a park, and a caf\'e.
    WanS2V (top) retains the seated pose despite the standing instruction, while MultiTalk (bottom) generates the pose change.}
    \label{fig:qual_motion}
\end{figure}

SceneShifter relies on the pretrained backbone to synthesize the requested subject motion.
In Fig.~\ref{fig:qual_motion}, the WanS2V-based result~\cite{gao2025wan} changes the background but retains the subject's seated pose, whereas the MultiTalk-based result~\cite{kong2025let} produces the requested standing motion.
Thus, successful scene control does not guarantee execution of motion instructions that the underlying model struggles to follow.

\subsection{Background Degradation in Short Scenes}
\begin{figure}[!ht]
    \centering
    \includegraphics[width=\textwidth]{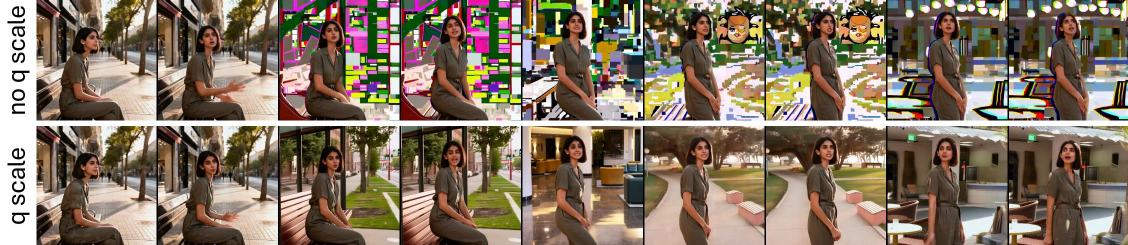}
    \caption{Effect of query scaling on short-scene generation with MultiTalk.
    The prompted scene sequence is an urban street, a bus stop, a hotel lobby lounge, a park, and a caf\'e.
    Without query scaling (top), later scenes exhibit severe background artifacts; query scaling (bottom) substantially improves their visual quality.}
    \label{fig:qual_qscale}
\end{figure}

Background quality can degrade when scene segments are much shorter than the base model's training clip length.
Fig.~\ref{fig:qual_qscale} shows severe background artifacts without query scaling.
Applying query scaling, following~\cite{jin2023training}, substantially reduces these artifacts in the illustrated example.
This mitigation improves short-scene generation but does not eliminate the dependence on the base model's capabilities.

\subsection{Severe Inter-Subject Occlusion}
\begin{figure}[!ht]
    \centering
    \includegraphics[width=\textwidth]{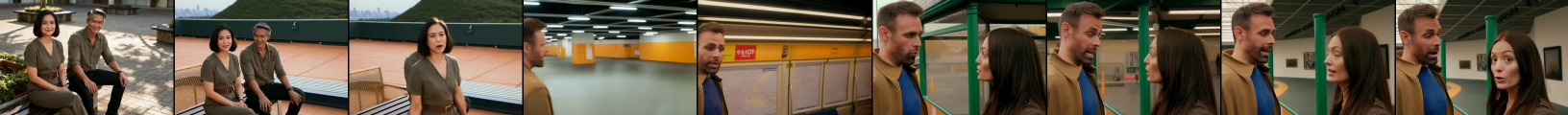}
    \caption{Failure case involving subject disappearance and re-entry.
    One of the two initial subjects disappears during the sequence, and their appearances differ substantially in later frames.
    The example illustrates the difficulty of maintaining subject identity when foreground continuity is interrupted.
    Frames are shown in temporal order.}
    \label{fig:qual_cyclic_failure}
\end{figure}

Under severe inter-subject occlusion, foreground masks can merge, causing a subject to disappear after a scene transition.
The resulting interruption of cross-scene foreground references can lead to appearance drift upon re-entry.
Fig.~\ref{fig:qual_cyclic_failure} illustrates subject disappearance and subsequent changes in appearance.
Preserving subject-specific references through heavy occlusion and temporary absence remains a direction for future work.

\subsection{Limitations of Automated Evaluation}
\label{sec:appendix_metric_limitations}
SceneShifterBench relies on automated scene-cut detection using PySceneDetect~\cite{pyscenedetect}, which can mistake camera motion or visual artifacts for scene transitions and miss gradual or subtle transitions.
These false positives and false negatives affect Scene Cut Success Rate and Scene Timing Accuracy.
Detection errors also influence Spatial Consistency, for which subject masks are compared at detected cuts, and Scene Semantics, for which detected cuts define scene segments and only videos satisfying the Scene Cut Success Rate criterion are included.
In future work, we will explore VLM-assisted scene detection to better distinguish semantic scene changes.

\section{Societal Impact}

SceneShifter can support creative video production, but may also be misused to create misleading depictions of individuals or harmful content.
Users could manipulate reference images to embed harmful contextual elements~\cite{jing2026cogmorph} or inject malicious prompts into video generation through jailbreak attacks~\cite{liu2025jailbreaking}.
Responsible deployment should prioritize consent, clear disclosure of synthetic content, and safeguards against malicious inputs~\cite{liang2026t2vshield}.

\end{document}